\documentclass[11pt]{article}

\usepackage{ACL}

\usepackage{times}
\usepackage{latexsym}

\usepackage{multirow}
\usepackage{xcolor}
\usepackage{tikz}
\usepackage{soul}
\definecolor{mistyrose}{HTML}{E9D8F7}
\definecolor{seafoam}{HTML}{C9E4CA}
\definecolor{warmcream}{HTML}{F2F2D7}

\usepackage[T1]{fontenc}

\usepackage[utf8]{inputenc}

\usepackage{microtype}

\usepackage{inconsolata}
\usepackage{graphicx}

\usepackage{booktabs}
\usepackage{array}

\newcolumntype{?}{!{\vrule width 1pt}}

\newcommand{\cut}[1]{}

\newcounter{tbsnr}
\newenvironment{tbs}
{\addtocounter{tbsnr}{1}\par\bigskip\noindent\fbox{\thetbsnr}
\hspace*{\fill}\begin{minipage}{7cm}\tt}
{\end{minipage}\hspace*{\fill}\bigskip}

\newcommand{\rb}[1]{\textcolor{blue}{\textbf{RB: #1}}}

\title{A Systematic Analysis of Large Language Models as Soft Reasoners:\\ The Case of Syllogistic Inferences}
\author{Leonardo Bertolazzi,\\
DISI, University of Trento\\
\texttt{leonardo.bertolazzi@unitn.it}\\\And
Albert Gatt,\\
ICS, Utrecht University\\
\texttt{a.gatt@uu.nl}\\\AND
Raffaella Bernardi\\
CIMeC and DISI, University of Trento\\
\texttt{raffaella.bernardi@unitn.it}}
\date{April 2024}

\begin{document}

\maketitle

\begin{abstract}
The reasoning abilities of Large Language Models (LLMs) are becoming a central focus of study in NLP. In this paper, we consider the case of syllogistic reasoning, an area of deductive reasoning studied extensively in logic and cognitive psychology. Previous research has shown that pre-trained LLMs exhibit reasoning biases, such as \textit{content effects}, avoid answering that \textit{no conclusion follows}, display human-like difficulties, and struggle with multi-step reasoning. We contribute to this research line by systematically investigating the effects of chain-of-thought reasoning, in-context learning (ICL), and supervised fine-tuning (SFT) on syllogistic reasoning, considering syllogisms with conclusions that support or violate world knowledge, as well as ones with multiple premises. Crucially, we go beyond the standard focus on accuracy, with an in-depth analysis of the conclusions generated by the models. Our results suggest that the behavior of pre-trained LLMs can be explained by heuristics studied in cognitive science and that both ICL and SFT improve model performance on valid inferences, although only the latter mitigates most reasoning biases without harming model consistency.\footnote{Our code and data are available at \texttt{\href{https://github.com/leobertolazzi/soft-syllogistic-reasoners.git}{https://github.com\\/leobertolazzi/soft-syllogistic-reasoners.git}}}
\end{abstract}

\section{Introduction}
\label{sec:intro}



\begin{figure}[t]
\begin{center}
  \includegraphics[width=1\linewidth]{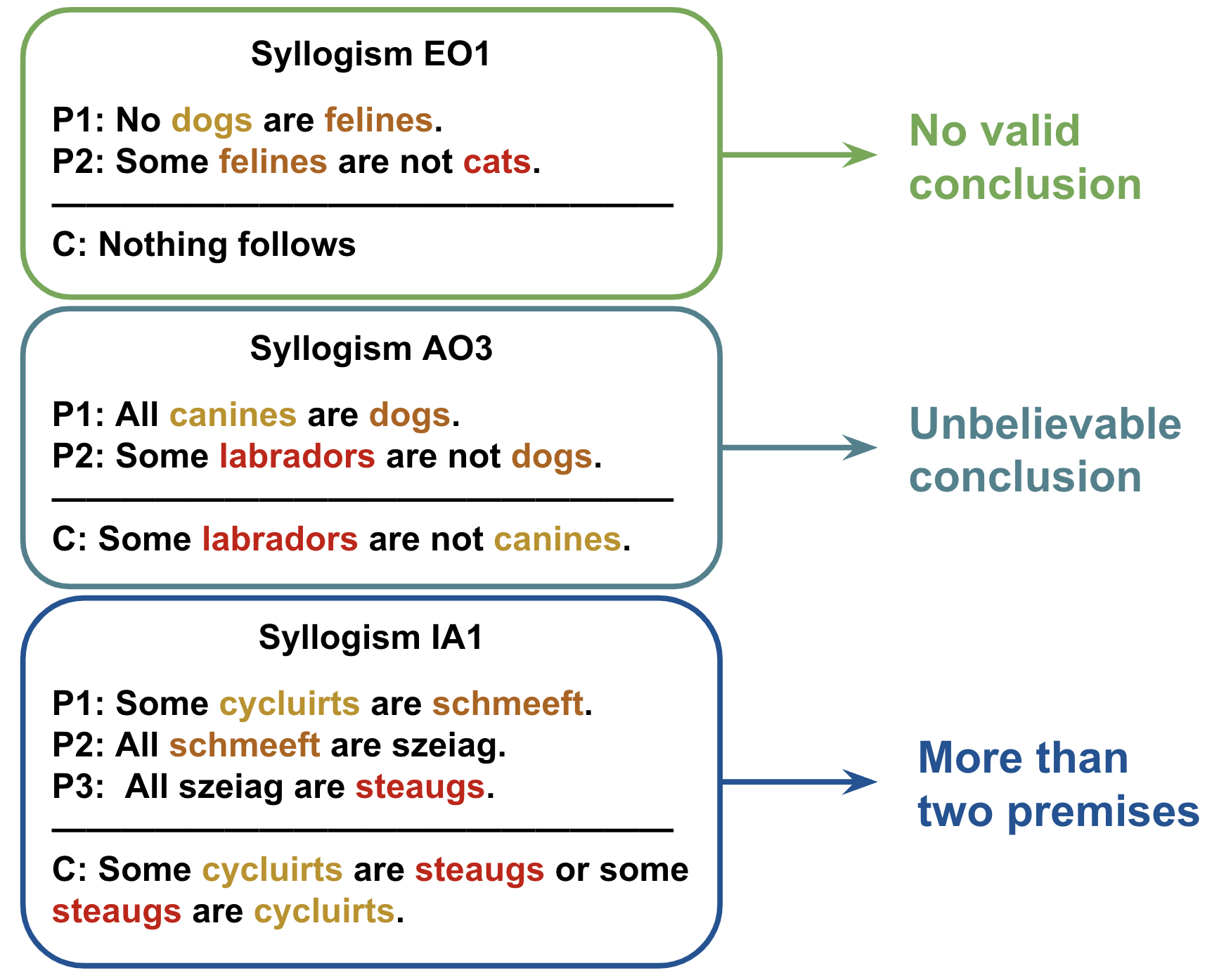}
\end{center}
\caption{LLMs have difficulty with invalid inferences (Top); suffer from content effects (Middle); and struggle with longer chains of premises (Bottom). What is behind such weaknesses? Can LLMs learn to use only the form to draw deductively valid conclusions?}\label{fig:bias}
\end{figure}

Soon after the first transformer-based models entered the NLP scene, \citet{ijcai2020p537} discovered that transformers are “soft reasoners”: they can be trained to emulate reasoning over natural language sentences. Since then, transformer-based language models have made enormous progress. A core question of today's AI research is whether Large Language Models (LLMs) have latently learned reasoning skills in addition to acquiring language proficiency.  If this is the case, such skills could be elicited through suitable prompting; if not,  other methods should be proposed to achieve computational models that can reason using language as the inference vehicle. One fundamental aspect of human reasoning is the ability to derive valid conclusions based on the form of the propositions involved, rather than their content~\cite{newell:huma72,newell:phys80}. This deductive reasoning skill is crucial in our everyday problem-solving and decision-making experiences, and its presence in LLMs is a topic of ongoing debate.


Recent research \citep{lampinen:lang23,eisape:syst23} shows that SOTA LLMs prompted with Chain-of-Thought (CoT) display humanlike reasoning biases in syllogistic reasoning; they have difficulties with the examples in Figure~\ref{fig:bias}, they (i) suffer from a content effect bias, favoring a conclusion compatible with world knowledge (that is, `believable'),
independently of whether it follows from the premises; (ii) struggle with syllogisms that humans also find hard; (iii) are not able to recognize invalid inferences -- premises from which no valid conclusion follows. Furthermore, LLMs struggle when multiple reasoning steps are required \cite{SaparovHe2023}, but \citet{PrOntoQAOOD} show that when prompted with CoT and in-context learning (ICL) examples, they manage to generalize to out-of-distribution logical rules.  We systematically bring these insights together, by comparing different learning strategies and by focusing on classical Aristotelian syllogisms, since they are an interesting test bed to understand whether an agent detaches form from meaning.

\cut{Recently, \citet{eisape:syst23} show that SOTA LLMs prompted with Chain-of-Thought (CoT)  display humanlike reasoning biases when processing the syllogistic fragment of logical inferences (see Figure \ref{fig:bias}), in that they (i) suffer from a content effect, producing a conclusion true in the world, independently of whether it follows from the premises; (ii) struggle with syllogisms that humans also find hard; (iii) are not able to recognize invalid inferences -- premises from which no valid conclusion follows. Comparisons among LLMs of different sizes, e.g.\ within the PaLM-2 family \cite{eisape:syst23}, among OpenAI models~\cite{SaparovHe2023} or across both types of models~\cite{lampinen:lang23}, show that the model size impacts reasoning performance, yet even larger LLMs  (above 7B parameters) suffer from human reasoning biases~\cite{eisape:syst23,lampinen:lang23,SaparovHe2023}, and have difficulty when multiple reasoning steps are required~\cite{SaparovHe2023}. Interestingly, LLMs prompted with CoT and in-context learning (ICL) examples manage to generalize to out-of-distribution logical rules~\cite{PrOntoQAOOD}.} 

 Given previous findings, we first study the performance of small- and medium-sized open-access models (Pythia and LLaMA) in Zero-Shot CoT (ZS-CoT) settings, relative to that of humans.
Subsequently, we examine the impact of both ICL and supervised fine-tuning (SFT) and consider how these approaches generalize to multi-step reasoning. 
This is achieved using carefully designed datasets of \textit{believable} and \textit{unbelievable} syllogisms, which respect or contradict world knowledge, respectively, and of syllogisms with 3- and 4-premises. Crucially, we go beyond the standard focus on accuracy: by looking at all conclusions the models generate we inspect their consistency and whether their answers are predicted by Heuristic Theories proposed in the cognitive science literature \cite{khem:theo12}. Our experiments reveal that:

\begin{itemize}
\item Open access models exhibit reasoning behaviours similar to humans,
a result that further strengthens findings recently published in~\citet{eisape:syst23}. They exhibit a reluctance to generate “nothing follows” for invalid syllogisms, and exhibit content effect biases when generating their answer, echoing results by~\citet{lampinen:lang23}; 
\item the answers generated by LLaMA-3 8B in the ZS-CoT setting are compatible with the predictions of  
the Atmosphere Heuristic Theory, suggesting that
it relies on the quantifier patterns in the premises. This would also explain 
pretrained LLMs' reluctance in generating “nothing follows”;
\item  though ICL boosts model performance on valid inferences, as~\citet{SaparovHe2023} suggest, it does not eliminate the content effect bias nor the difficulty in handling invalid syllogisms. It also increases inconsistency, whereby models generate contradictory conclusions;
\item SFT is effective for both small- and medium-sized models, and lets the latter reach near-perfect performance while remaining consistent. It has to be seen how much this ability is transferred to the real-life scenarios a "soft reasoner" should handle.  
\end{itemize}

\section{Related Work}
\label{sec:rw}

\begin{figure*}[t]
\begin{tabular}{ccc}
\emph{moods} & \emph{figures} & schema: AE2\\[2mm]
\begin{tabular}{l|l}
affirmative & negative\\
\textbf{A}: All $a$ are $b$ & \textbf{E}: No $a$ are $b$\\
\textbf{I}: Some $a$ are $b$  & \textbf{O}: Some $a$ are not $b$
\end{tabular}
&
\begin{tabular}{lllll}
& 1 & 2 & 3 & 4\\\hline
P1: & a-b & b-a & a-b & b-a\\ 
P2: & b-c & c-b & c-b & b-c
\end{tabular}
&
\begin{tabular}{ll}
P1: & All $b$ are $a$ (A)\\
P2: &  No $c$ are $b$ (E)\\
C: & Some $a$ are not $c$
\end{tabular}
\end{tabular}

\caption{\textbf{The building blocks}: Moods (A, E, I, O) and figures (1-4). Their combination determines the conclusion, as illustrated by the AE2 schema.}\label{fig:rules}
\end{figure*}


Before the recent surge of general-purpose LLMs, reasoning in transformer language models was primarily studied in the field of Natural Language Inference (NLI), a long-standing challenge within the computational linguistic comunity~\cite{giampiccolo:2007-rte}. The most recent research mainly utilized encoder-only models like BERT~\cite{devlin:2019}, RoBERTa~\cite{liu:2019}, and XLNet~\cite{yang:2019}. These models approached NLI as a classification task, using datasets such as SNLI~\cite{bowman:2015} and MNLI~\cite{williams:2018} for fine-tuning. The literature in this area focused on both inductive and deductive inferences, with deductive inferences further categorized into formally valid and materially valid inferences (for a recent survey see \citealt{gubelmann:2024}).

With the emergence of larger and more powerful decoder-only LLMs, many NLP tasks no longer required specific architectures or training. Instead, a single model could perform most tasks. This shift led to increased interest in studying more abstract reasoning skills of LLMs~\cite{huang-chang-2023-towards,qiao-etal-2023-reasoning,Mondorf:beyo24}. It has been established that Chain-of-Thought (CoT) prompting elicits some reasoning capability~\cite{wei:chain22,kjim:llm22} and that in-context learning (ICL) further boosts such skills~\cite{huang-chang-2023-towards}. In this paper, we focus on syllogisms because they are an interesting test-bed to study the interleaved role of content and form in deriving conclusions, and have been extensively studied in the cognitive science literature.


\citet{liu2023glore} introduced an assembled General Logical Reasoning Evaluation benchmark (GLoRE) and compared ChatGPT, GPT-4, and humans
against open access LLMs, such as LLaMA~\cite{touvran:llama23}. They show that 
instruction-tuning, supervised fine-tuning, in-context learning, and voting
techniques increase LLaMA's task accuracy. They also raise the possibility that
models rely on superficial patterns in solving logical reasoning tasks.  With this observation in mind, we first consider models' accuracy and correlation with human performance, then, taking inspiration from
cognitive studies about human heuristic reasoning models~\cite{khem:theo12}, we run a fine-grained analysis of their answers.


The two papers that are most related to our work
are~\citet{lampinen:lang23} and~\citet{eisape:syst23}.
\citet{lampinen:lang23} challenge proprietary LLMs 
with discriminative reasoning tasks, including the classification of syllogistic inferences as valid or invalid,
and show that larger models are more logical than smaller ones, and LLMs suffer from the content effect bias. We extend this result by looking at LLMs as generative models. \citet{eisape:syst23} evaluate the PaLM 2 family~\cite{google:palm} on a generative version of the syllogism task; specifically, they use Zero-Shot CoT prompting~\cite{wei:chain22,kjim:llm22}
to select which conclusion, among a list of candidates, is
the one that logically follows from a pair of premises. 
They show that even the larger models make mistakes that mirror human reasoning biases. Moreover, they show that LLMs have difficulty with invalid syllogisms: they rarely generate "nothing follows"  even when it is presented as one of the multiple choices.  We extend this analysis of ZS CoT models by systematically investigating the effects of CoT reasoning, ICL, and supervised fine-tuning (SFT) on syllogistic reasoning, considering syllogisms with conclusions that support or violate world knowledge and with multiple premises. 

 
\citet{PrOntoQAOOD} show that models with CoT and ICL generalize to rules not seen inside their context but fail in generalizing to longer proofs. 
We advance our understanding of the role of ICL by focusing on the interplay between content and form and carrying out in-depth analysis of the conclusions.

In summary, previous contributions have identified various important dimensions and 
separately addressed the impact of Zero Shot CoT, ICL, and SFT. In this paper, we systematically address these dimensions, compare model and human performance, carry out an in-depth analysis of outputs, an analysis missing in the current literature.


\section{Syllogisms}

Syllogisms consist of two premises and a conclusion that share the
structure ``Quantifier X are Y''. As summarized in Figure~\ref{fig:rules} (left) such statements can be in
four \emph{moods}: either affirmative or negative, and either
universal or existential based on the type of quantifier (\emph{“all”}, \emph{“some”} or \emph{“some \ldots not”} and \emph{“no”}); they are abbreviated using
letters -- affirmatives as ``A'' and ``I'', negatives as ``E'' and ``O'' from
the Latin ``AffIrmo'' and ``nEgO'', respectively. The order in
which the nouns occur within the premises defines the \emph{figure}; there are four
possible figures marked by numbers (1-4). 
Only one noun (represented in Figure~\ref{fig:rules} by the $b$ variable) is shared between the premises, the relation holding between the other two nouns (the $a$ and $c$ variables), if any, is derived from the premises. Hence, syllogisms consist of a chain of predications linking the terms of the conclusion.

Such inference can be defined by the combination of moods and figures, which are referred to as "schemas" and abbreviated by naming the premise moods and the figure number. Figure~\ref{fig:rules} exemplifies the AE2 schema. 
Interestingly, schemas can be composed to obtain longer inferences (longer chains) and can be applied recursively, requiring multi-hop reasoning~\cite{guzman:test24}. For instance, to infer a conclusion for the premises in Figure~\ref{fig:bias} (bottom), one should first derive an intermediate conclusion from two premises, e.g., P1 and P2, and then combine the intermediate conclusion with the remaining premise to derive the final conclusion. This gives rise to multiple paths since more intermediate conclusions are possible.


Since each premise can be in one of the four moods, there are 64
distinct pairs of premises (16 moods x 4 figures).
We follow the psychology literature~\cite{khem:theo12} and consider 37 of these
combinations as invalid, while the remaining 27 lead to at least one 
conclusion (for the full list see Appendix \ref{app:data}). 




Humans exhibit a \textit{content effect bias}, namely the
tendency to consider an inference valid when the conclusion is
believable and invalid otherwise (e.g.,~\citealp{evans:memo83}).
In their meta-review, \citet{khem:theo12} aggregate results from six independent studies run with students of different levels (from high school to university) and show that the overall patterns of responses are robust and reliable; for humans, some schemas are  easier than others, e.g., on the AI2 and the AE2 schemas, they obtain 90\% and 1\% accuracy, respectively. Finally, humans seem to process valid and invalid syllogisms differently and have difficulty deciding that no valid conclusion follows~\cite{ragni:when19}. 

\begin{figure*}[t]
\begin{center}
  \includegraphics[width=0.95\linewidth]{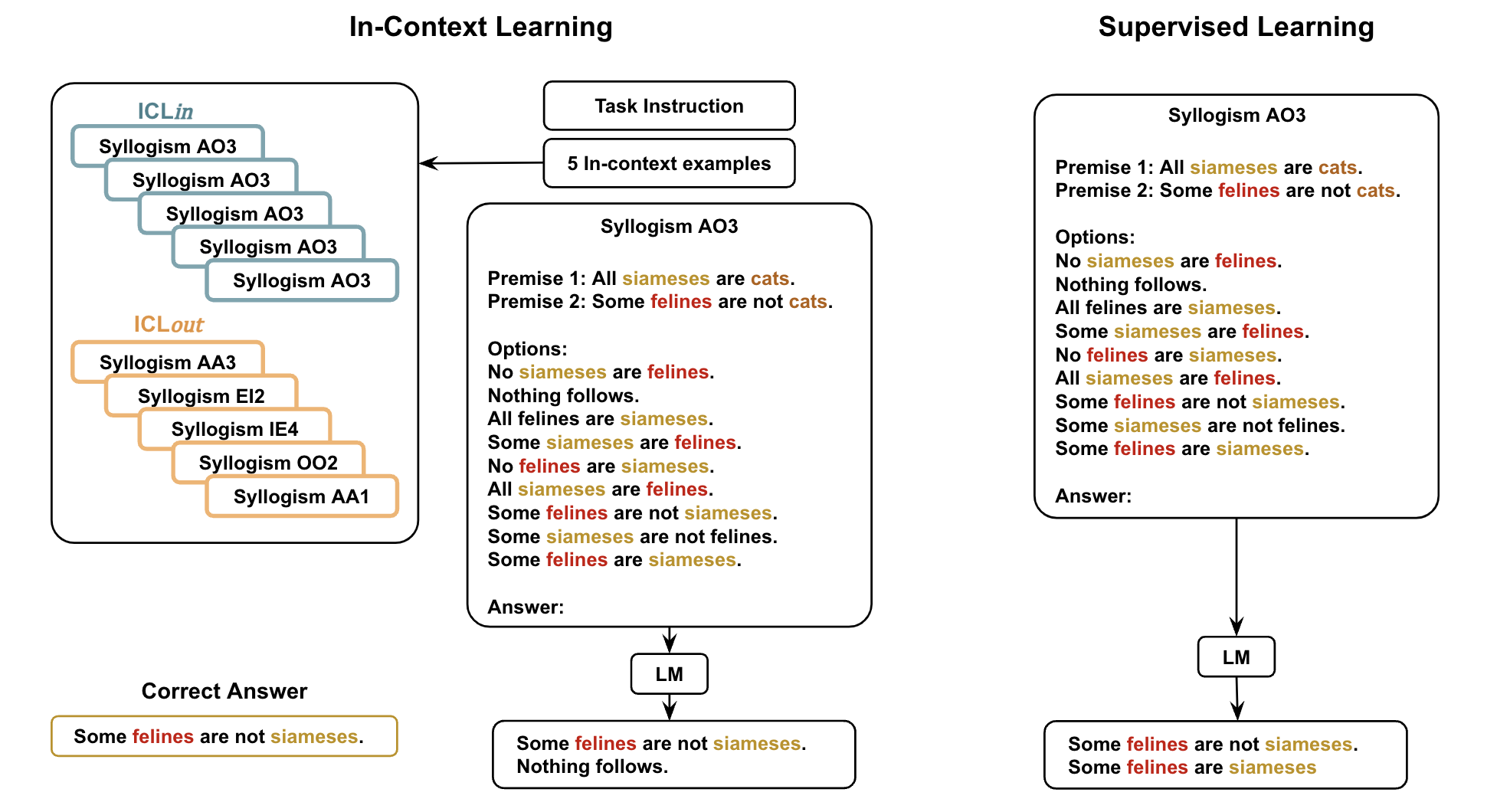}\\
\end{center}
\caption{
\textbf{Multiple-choice Task} The model is given the premises and nine possible conclusions, and has to generate the correct one(s). $\textrm{ICL}_{out}$ is given in-context examples of different schemas than the one of the test example, while $\textrm{ICL}_{in}$ receives in-context examples of the same schema. The supervised fine-tuned model is trained on all schemas.}\label{fig:task}
\end{figure*}

\section{Learning Strategies}

We aim to understand whether LLMs are equipped with latent reasoning abilities to extract syllogistic schemas from given examples or if they could achieve such ability through supervised learning. To this end, we compare in-context and supervised learning.
Given the premises, we frame the problem as a multiple-choice task (see Section \ref{sec:data}).




\paragraph{Baseline}   We take as baselines the Zero-shot CoT (\textbf{ZS-CoT}) setting \citep{kjim:llm22}. Zero-shot CoT is a two-stage prompting method that provides no task examples to a model. In the first stage, a model is given a task instruction and prompted to think through a series of reasoning steps before generating a response - typically by adding the phrase “Let's think step by step” before its response. In the second stage, the model is asked to provide a final answer based on its own reasoning chain. We take Pythia and LLaMA pre-trained models, and compare them in ZS-CoT looking at the effect of model size and, when possible, of undergoing a further instruction-tuning phase.\footnote{All the models we used are available in the \href{https://huggingface.co/}{HuggingFace model hub}. The instruction tuned version of Pythia-1.4B we used can be found under the id \texttt{lambdalabs/pythia-1.4b-deduped-synthetic-instruct}}

\paragraph{In-Context Learning} In this setting, there is no further training, hence no weight changes. A model is presented with $k$  examples of the multiple-choice task. 
It has been claimed that the diversity of the ICL examples increases the model's performance~\cite{levy-etal-2023-diverse}. Hence, as illustrated in Figure~\ref{fig:task}, we
evaluate the setting in which the test example belongs to the same schema as the $k$ randomly chosen in-context examples  (\textbf{ICL$_{in}$}), or belongs to a different one (\textbf{ICL$_{out}$}); we set $k=5$. 

\paragraph{Supervised Fine-Tuning (SFT)} We train models for $n$ epochs on the task using standard cross-entropy loss and next-token prediction objective. At each epoch, the model sees only one instance of each schema; the instances, viz.\ the terms replacing the $a$, $b$, $c$ variables, change through the epochs.




\section{Task and Datasets}
\label{sec:data}

\paragraph{Task}  
Following~\citet{ragni:when19,eisape:syst23}, we use a
multiple-choice setting. As illustrated in Figure~\ref{fig:task}, for a given syllogistic schema, the model is given two 
premises followed by nine possible conclusions. These nine conclusions include all the possible relations between the terms $a$ and $c$ ($Aac$, $Iac$, $Eac$, $Oac$, $Aca$, $Ica$, $Eca$, $Oca$), plus the string ``Nothing follows'', which has to be selected when from the two premises none of the other options can be drawn. These possible conclusions are always presented to the model in random order. To succeed in the task, the model has to generate at least one of the correct conclusions.\footnote{Depending on the schema, there can be from 1 to 4 correct conclusions.} Below, we will describe how we built the training data, the ICL examples, and the various test sets.

\paragraph{Training sets}  We created our training sets by replacing variables with pseudo-words. We use 4k
pseudo-words\footnote{We generated a vocabulary of 4k pseudo-words using the \href{https://pypi.org/project/gibberish/}{\texttt{gibberish}} Python library.} to produce inferences that preserved the formal structure of the original data points but have nonsensical content. We use these data points as ICL examples and as training data for SFT. For the latter, to 
select the best model, we use a development set, containing one instance of each schema (64 data points) built with a vocabulary of 1k pseudo-words not seen during training. Further details on the training are reported in Appendix \ref{app:impl}.

\paragraph{Test sets}
To study the content-effect bias we build believable and unbelievable syllogisms. Similarly to~\citet{lampinen:lang23}, we use triples of
terms that are ontologically related: we manually generated 10 triples where terms are in a taxonomical relationship 
(e.g. \texttt{siameses}, \texttt{cats}, \texttt{felines}) and substituted the $a$, $b$, and $c$ variables of the schema with these triples, obtaining premises that are true in the real world. We refer to this set as \emph{believable} syllogisms. An example is provided in Figure~\ref{fig:task}. 
Since the content effect relies on the plausibility of the conclusion, we generated \emph{unbelievable} syllogisms only for the valid schemas. We used the same 30 terms but made sure that the correct conclusions violate real-world knowledge,
as illustrated by the example in Figure~\ref{fig:bias} (middle). 
For the believable syllogisms, we generated test sets consisting of 640 data points each (64 schemas instantiated with 10 triples of terms); for the unbelievable ones, we generated 270 data points (27 valid schemas instantiated with 10 triples of terms).



To check the models' robustness to inference length, we create inferences with more than two premises whose solution requires applying a chain of syllogistic schemas.
To build these \emph{longer inferences}, we exploited the property of transitivity replacing an A-premise ($Aab$) with a chain $Aac_1,Ac_1c_2, ... , Ac_{n-1}c_n, Ac_nb$. By setting $n=2$ and $n=3$, we obtain inferences with 3 and 4 premises, respectively.
Out of the 64 schemas, 28 have at least one A-premise, hence, we obtained 280 data points both for 3- and 4-premises settings (28 schemas instantiated with 10 different triples of pseudo-words not present in the training and development sets). Further details on the datasets construction are in Appendix~\ref{app:data}.


\section{Results based on task-accuracy}
\label{sec:expres}

In this section, first, we look into ZS CoT accuracy, then we compare different learning strategies to address the question of whether LLMs can learn to rely only on the form to generate a conclusion, an issue which has not received a systematic analysis yet.





\subsection{Zero-shot CoT Setting}
Human accuracy is 44.63\% on valid inferences and around 40.97\% on invalid ones~\cite{khem:theo12}. 
As Table~\ref{tab:zeroshot} shows, on valid syllogisms, all versions of  Pythia are well below human performance, 
while LLaMA-2 reaches it and the recently released LLaMA-3 even surpasses humans. However, none of the models reach human accuracy on invalid inferences, the closest being LLaMA-3 70B instruction-tuned (32.50\%).
Instruction tuning is beneficial for invalid syllogisms for all models and all sizes, but at the price of decreasing performance on valid syllogisms. All models have a high correlation\footnote{See Appendix \ref{app:exp} for a more detailed explanation of the correlation results.} with human score patterns, with LLaMA-3 70B models having the highest. When the correct conclusion is false in the real-world (unbelievable set), most of the models have a drop in accuracy compared with valid believable inferences. The results reported so far (Table \ref{tab:zeroshot}) are in line with recent findings by~\citet{eisape:syst23} on LLMs correlation with humans and their reluctance to generate “nothing follows”, and extend the results on content effects from the discriminative task used by~\citet{lampinen:lang23} to the generative one we consider. 

\begin{table*}
\begin{center}
    \begin{tabular}{lcccccc}
    \toprule
    \multirow{2}{*}{\textbf{Model}} & \multicolumn{3}{c}{\textbf{Believable}} & \textbf{Unbelievable} & \textbf{Content Effect} & \textbf{Correlation}\\
\cmidrule(lr){2-4} \cmidrule(lr){5-5} \cmidrule(lr){6-6} \cmidrule(lr){7-7}
& Acc & Invalid & Valid & Valid & Difference & Spearman $\rho$\\\midrule
Pythia-160M & 0.16 & 0.00 & 0.37 & 0.00 & -100.00\%* & 0.10 \\
Pythia-410M & 9.53 & 0.00 & 22.59 &  19.63 & -13.10\% & 0.73 \\
Pythia-1.4B & 14.22 & 1.94 & 31.11 & 33.33  & +7.14\% & 0.71 \\
Pythia-1.4B-inst & 17.03 & 15.28 & 19.26 & 15.93 & -17.29\%* & 0.45 \\\midrule
LLaMA-2 7B & 24.38 & 3.06 & 53.70 &  47.78 & -11.02\% & 0.56 \\
LLaMA-2-inst 7B & 19.69 & 10.83 & 32.22 &  19.26 & -40.22\% & 0.36 \\
LLaMA-2 13B & 22.81 & 0.28 & 53.70 &  40.37 & -24.82\% & 0.74 \\
LLaMA-2-inst 13B & 22.34 & 8.06 & 42.22  & 35.93 & -14.19\%* & 0.65\\\midrule
LLaMa-3 8B & 30.31 & 0.83 & 70.74 &  52.59 & -10.44\% & 0.48 \\
LLaMA-3-inst 8B & 50.00 & 30.28 & 78.15  & 64.81 & -17.07\% & 0.30 \\
LLaMA-3 70B & 30.63 & 10.28 & 58.89 &  54.07 & -8.18\% & 0.80 \\
LLaMA-3-inst 70B & 41.25 & 32.50 & 54.44 & 57.41 & +5.46\% & 0.87 \\\bottomrule
        \end{tabular}
      \caption{\textbf{Zero-shot CoT (ZS-CoT)}: Humans reach an accuracy of 44.63\% on valid and of 40.97\%  on invalid syllogisms~\cite{khem:theo12}. We compare model performance by reporting the overall accuracy on believable syllogisms as well as its breakdown into valid vs.\ invalid schemas: Pythia models are well below human scores, while LLaMA models reach or outperform humans on valid syllogisms but are unable to recognize invalid inferences. This weakness is mitigated by instruction tuning but comes at the price of decreasing performance on valid syllogisms. All LLaMA models except 70B instruction tuned, suffer from content effect bias (the performance loss from the believable to the unbelievable set on valid syllogisms). $^*$ marks when the performance loss is \textit{not} statistically significant based on $\chi^2$’s Test. On the right, Spearman $\rho$ correlation coefficient between the accuracy of humans and models on the valid schemas of the believable set. The human accuracies used to compute the coefficient are based on \citet{khem:theo12}.}\label{tab:zeroshot}
    \end{center}
\end{table*}

\begin{table*}
\begin{center}
\begin{small}
    \begin{tabular}{llcccccc}
       \toprule
  &  & \multicolumn{3}{c}{\textbf{Believable}} & \textbf{Unbelievable} & \textbf{Content Effect} & \textbf{Correlation}\\
\cmidrule(lr){3-5} \cmidrule(lr){6-6} \cmidrule(lr){7-7} \cmidrule(lr){8-8}
& & Acc & Invalid & Valid & Valid & Difference & Spearman $\rho$ \\\midrule
\multirow{4}{*}{\rotatebox{90}{\sc Pythia}} & ZS-CoT & 14.22 & 1.94 & 31.11 & 32.96 & 7.14\% & 0.71 \\
& ICL$_{out}$ &  16.67 ($\pm 1.1$) & 1.57 ($\pm 0.7$) &  37.41 ($\pm 3.53$) & 32.72 ($\pm 1.67$) & -12.54\% & 0.63 \\
& ICL$_{in}$ & 22.55 ($\pm 0.39$) & 7.13 ($\pm 1.4$) &  43.07 ($\pm 0.98$) &  36.91 ($\pm 6.59$) & -15.54\% & 0.61 \\
& SFT &  92.97 ($\pm 4.51$) & 93.98 ($\pm 5.14$) &  91.48 ($\pm 4.27$) &  90.74 ($\pm 4.19$) & -2.30\% & 0.16 \\\midrule 
\multirow{4}{*}{\rotatebox{90}{\sc LLaMa}} & ZS-CoT  & 30.31 & 0.83 & 70.74 & 52.59 & -25.66\% & 0.74 \\
& ICL$_{out}$ & 29.58 ($\pm 1.11$) & 1.85 ($\pm 0.42$) & 67.53 ($\pm 2.88$)  & 62.96 ($\pm 2.06$) & -6.77\% & 0.55 \\ 
& ICL$_{in}$ & 33.39  ($\pm 0.55$) & 1.39 ($\pm 1.00$) & 77.16 ($\pm 2.38$) &  73.09 ($\pm 1.19 $) & -5.27\% & 0.74 \\
& SFT & 96.25 ($\pm$ 2.48) & 97.13 ($\pm$ 3.83) & 94.94 ($\pm$ 7.18) &  99.38 ($\pm$ 0.77)  &  4,68\% & -0.07 \\\bottomrule
    \end{tabular} 
\end{small}
    \caption{\textbf{Learning strategies}: Models: Pythia 1.4B, LLaMA-3 8B. Among ICL settings, ICL$_{in}$ outperforms ICL$_{out}$ in generating valid conclusions; SFT achieves higher results than ICL, it is stable across models, performing well on both valid and invalid syllogisms, and lets the model overcome the content bias. For both models, only SFT overcomes the difficulty of generating the “nothing follows” answer. Results are the average over three runs, changing seeds and training data/in-context examples. The standard deviation is in parentheses. Right: Spearman $\rho$ correlation coefficient between models' accuracies and those obtained by humans on valid syllogisms from the \textit{believable} set. The human accuracies used to compute the coefficient are based on \citet{khem:theo12}.}\label{tab:learning}
     \end{center} 
\end{table*}

\begin{table*}[t]
    \centering
\begin{tabular}{llccc}
\toprule
\multicolumn{2}{l}{} & \textbf{2 premises} &  \textbf{3 premises} & \textbf{4 premises}\\\midrule
\multirow{4}{*}{\rotatebox{90}{\sc Pythia} }& ZS-CoT & 18.21  & 11.07 & 7.86 \\
&  ICL$_{out}$ & 22.14 ($\pm 1.99$) & 19.05 ($\pm 1.09$)  & 20.12 ($\pm 2.18$)  \\
& ICL$_{in}$ & 30.71 ($\pm 3.11$)  & 29.88 ($\pm 1.49$)  & 25.48 ($\pm 4.61$) \\
& SFT & 83.33 ($\pm 13.41$) & 67.14 ($\pm 4.06$)  & 53.57 ($\pm 3.98$)  \\\midrule 
\multirow{4}{*}{\rotatebox{90}{\sc LLaMA}} & ZS-CoT & 44.64 & 39.64 &  31.79\\
& ICL$_{out}$ & 39.29 ($\pm 1.64$) & 40.48 ($\pm 1.76$)  & 37.98 ($\pm 0.41 $) \\
& ICL$_{in}$ & 53.69 ($\pm 1.03$)  & 46.90 ($\pm 1.80$)  & 49.52 ($\pm 2.03$) \\
& SFT &  99.05 ($\pm 0.9$) & 94.17 ($\pm 2.15$)  & 85.36 ($\pm 6.23$)  \\\bottomrule
\end{tabular}
\caption{\textbf{Unseen number of premises}: Models: Pythia 1.4B, LLaMA-3 8B. Premises contain pseudo-words in all three test sets. The 2-premise set contains 10 instances of the 28 schemas with at least one A premise. The other sets are obtained by replacing 1 A premise (Aab) with a chain of 2 A premises (Aac, Acb, 3-premise set) or 3 A premises (Aac, Acd, Adb, 4-premise set). The ZS-CoT's performance decreases with a longer chain of premises, showing the task becomes harder.  ICL$_{out}$ is more robust than ICL$_{in}$  but SFT remains the best-performing setting on all sets despite the increased number of premises.}\label{tab:length}
\end{table*}

\subsection{Learning strategies performance}
We take the largest version of Pythia (Pythia-1.4B) and the smallest version of LLaMA (LLaMa-3 8B), both without instruction tuning, to evaluate them in the ICL and SFT settings.

\paragraph{Human and AI biases}
As illustrated in Table~\ref{tab:learning}, ICL does not help models recognize when no valid conclusion follows. This is particularly striking, given that in ICL$_{in}$, the in-context examples are of the same schema as the test examples, hence the model fails even to copy the “nothing follows” answer.\footnote{See an example in Appendix \ref{app:examples}.} This weakness is overcome by SFT: both Pythia and LLaMA 
are at or near ceiling 
on both invalid and valid inferences.
Interestingly, models only overcome content effect biases in the SFT setting, which suggests that hindering a reliance on lexical content (e.g. through pseudo-words) is effective in
forcing models to draw conclusions based on form rather than content.\footnote{We have run an in-depth analysis of the content effect bias by inspecting whether models prefer generating believable conclusions when the unbelievable ones are correct (see Appendix \ref{app:exp}). The analysis confirms the results reported here.} 

\paragraph{Unseen number of premises} Handling a higher number of premises is harder, as illustrated in Table~\ref{tab:length} by the decreased performance of the models tested in the ZS CoT setting. As explained in Section~\ref{sec:data}, here we are focusing on 28 schemas (those having an A-premise) --  a total of 280 data points with 2, 3 or 4
premises containing pseudo-words instead of nouns. We see that indeed both models are more robust when given diverse in-context examples (ICL$_{out}$ has a lower drop than ICL$_{in}$ with a higher number of premises.) Overall, SFT has a higher drop than ICL, but it still has a much higher accuracy than ICL both for 3- and 4-premises. This result aligns with previous research, which shows that models trained or fine-tuned to determine whether an inference can be drawn from a given set of premises struggle to extrapolate their abilities when faced with a longer set of premises \citep{ijcai2020p537, guzman:test24}.

\section{Beyond task accuracy}
\label{sec:qual}

So far our evaluation has focused on whether there is at least one correct conclusion among those generated. In this section, we identify what is behind the weaknesses of models, by analyzing in depth \emph{all conclusions} generated.

\begin{table*}[t]
    \centering
\begin{tabular}{clcc?ccc}
\toprule
\multicolumn{2}{l}{} & \multicolumn{2}{c?}{\textbf{Consistency}} & \multicolumn{2}{c}{\textbf{Predicted Mistakes}} &  \textbf{Predicted Correct}\\
 \cmidrule(lr){3-4} \cmidrule(lr){5-6} \cmidrule(lr){7-7}
 \multicolumn{2}{l}{} & \% AO, EI, NVC+ & \% NVC+ & Invalid & Valid & Valid \\\midrule
\multirow{4}{*}{\rotatebox{90}{\sc Pythia}} & ZS-CoT & 18.91 & 13.93 & 24.37 & 12.04 & 70.00 \\ 
& ICL$_{out}$ & 28.12 & 28.12 & 17.60 & 8.44 & 61.22 \\ 
& ICL$_{in}$ & 28.28 & 22.48 & 12.57 & 9.45 & 61.36 \\
& SFT & 73.91 & 70.47 & 96.77 & 18.31 & 54.04 \\ \midrule 
\multirow{4}{*}{\rotatebox{90}{\sc LLaMA}} & ZS-CoT & 1.09&  0.78 & 82.97 & 62.79 & 92.31 \\
& ICL$_{out}$ & 1.72 & 1.09 & 62.22 & 25.00 & 61.54\\
& ICL$_{in}$ & 3.12 & 2.23 & 47.88 & 31.51 & 63.50 \\
& SFT & 0.31 & 0.00 & 0.00 & 12.50 & 65.12 \\\bottomrule 
\end{tabular}
\caption{\textbf{Model consistency} (Left): \% of contradictory answers. ICL makes all models more contradictory. SFT LLaMA generates almost no contradictory answers (only 0.31\% of its answers are  {\bf AO}: a universal affirmative and an existential negative, or {\bf EI}: universal negative and existential positive; while it never generates “nothing follows” together with other conclusions (NVC+). Instead, SFT causes Pythia to correctly answer “nothing follows”, but at the price of increasing its inconsistency, since it accompanies this answer with other conclusions too. \textbf{Heuristic Theory: Atmosphere} (Right) Proportion of models' conclusions predicted by the Atmosphere Theory (AT). AT is a good model of LLaMA ZS-CoT: it predicts its mistakes both in invalid (82.97\%) and valid inferences (62.79\%) as well as its correct conclusions (92.31\%) for valid schema; this would suggest that the model relies on the moods of the premises to generate the conclusions, which would explain the low percentage of “nothing follows”.}\label{tab:unbe-cont}
\end{table*}

To inspect model consistency, we focus on the believable syllogisms and analyze whether models generate contradictory conclusions.\footnote{An additional analysis of the completeness of the answers generated is provided in Appendix \ref{app:inc}.}
Among the nine possible conclusions, the contradictory pairs are: a) Universal affirmative and existential negative, viz.\ “All $a$ are $b$”  and “Some $a$ are not $b$” (AO); b) A universal negative and existential positive, viz. , “No $a$ are $b$” and “Some $a$ are $b$” (EI); c) “Nothing follows” and any of the other possible conclusion (NVC+), as illustrated in Figure~\ref{fig:task} with the example of the ICL LM's answers.  Table~\ref{tab:unbe-cont} (left) reports the percentage of contradictory answers, highlighting a rather different behavior of the two models. While LLaMA-3 8B ZS-CoT hardly contradicts itself (1.09), Pythia does it much more (18.91).  For both models, ICL increases the number of contradictory answers, with ICL$_{in}$ damaging model consistency even more than ICL$_{out}$. SFT reduces the contradictory answers of LLaMA even further, while Pythia still generates
73.91\% contradictory answers. Interestingly, SFT causes Pythia to overgenerate "nothing follows": of the 73.91\% contradictions, 70.47\% are due to the presence of “nothing follows” (NVC+).

As a final analysis, we check whether the models have learned some heuristics. By inspecting human responses across a variety of experiments, \citet{khem:theo12} analyze the heuristic theories proposed as potential theories of human reasoning, concluding 
that none of these theories are good models of human reasoning: a good model should predict both the correct and the wrong conclusions an agent draws from the premises. We use these theories as a tool to further inspect models. We focus on the Atmosphere theory since it turns out to be the most insightful.\footnote{See Appendix \ref{app:heur} for the detailed predictions and the complete results of the other heuristic theories.} The theory claims that “reasoners tend to accept a conclusion that fits the mood of the premise”. In other words, the conclusion is drawn considering only the quantifiers in the premises. Among all the correct conclusions for all syllogisms, the Atmosphere theory can predict 62.50\% of them on valid syllogisms and 0\% on invalid schemas.

%
 As reported in Table~\ref{tab:unbe-cont} (right), the Atmosphere Theory predicts 92\% of the correct valid conclusions generated by LLaMA ZS-CoT as well as its errors in invalid (82.97\%) and valid (62.79\%) schemas. Since the Atmosphere Theory fits the model predictions very well, this could imply that the model relies on the moods of the premises to draw conclusions, which would explain why it rarely generates “nothing follows”. 

In all our analysis we noticed distinct patterns for the two SFT models, namely Pythia struggles more with an unseen number of premises (Table \ref{tab:length}), becomes inconsistent at the price of increasing its accuracy (Table \ref{tab:atmosphere_pred}, Left), and most of its mistakes on invalid syllogisms are predicted by the Atmosphere Heuristic (Table \ref{tab:atmosphere_pred}, Rigth). We believe these differences between LLaMA and Pythia highlight the impact of the number of parameters and the amount of data used during the pretraining phase on the models’ performance during fine-tuning.

Although the results show that LLaMA SFT does not adopt a superficial strategy like the Atmosphere Heuristic and is more consistent, we cannot exclude that these results may still be driven by a more sophisticated form of pattern matching rather than genuine learning of the correct inference rules. This hypothesis is supported by the results on previously unseen numbers of premises (see Table \ref{tab:length}), where SFT models exhibit the greatest performance drop compared to other strategies.

\section{Conclusion}

Models that can reason using language as the inference vehicle, or “soft theorem provers” in the words of~\citet{ijcai2020p537}, have been the Holy Grail of AI since its inception. 
Now that LLMs have achieved amazing language proficiency skills, the community is investigating their reasoning ability. We contribute to these efforts by focusing on syllogistic inferences.
We systematically compared ZS-CoT, ICL, and SFT  and analyzed the answers generated by the models in depth.
To our knowledge, we are the first to use heuristic theories, such as the Atmosphere Theory, to study LLMs. This heuristic turns out to be a very good predictor for LLaMA-3 8B ZS-CoT; therefore, 
we conjecture that the reason Zero-Shot LLMs rarely generate “nothing follows” is that they use the quantifiers in the premises as cues. 


The merit of ICL has been advocated by~\citet{SaparovHe2023}. By going beyond task accuracy and inspecting all the conclusions models generate, we show that while ICL boosts model performance on valid inferences, it does not eliminate content effects or the difficulty in handling invalid syllogisms. Most importantly, it increases model inconsistency. On the other hand, fine-tuning the model on next-word prediction is effective for both the small- and medium-sized models we evaluated and allows the latter to approach 
ceiling performance while remaining consistent.  

\citet{langmodelsopinion} argue that language and reasoning are
dissociated, and to be enhanced with the latter, LMs should go
through a different learning paradigm or be equipped with additional
modules. 
Our
results suggest that while content-based reasoning can be learned from pre-training, form-based reasoning needs appropriate training.  We leave for the future the question of whether LLMs after SFT on a controlled dataset, as the one we use, transfer the acquired reasoning ability to real-life scenarios with linguistically richer expressions.





\section{Limitations} 
We limited our analysis to two families of models - Pythia and LLaMA 2, 3. However, a more general investigation on more families of open and closed models would lead to a better understanding of how ZS-CoT, ICL, and SFT would affect LLMs abilities in syllogistic inferences. In addition, due to the limitations of available computational resources, we limited the analysis of ICL and SFT to small and medium-sized open models, excluding larger models (LLaMA-3 70B).

The generative approach used in our work enabled models to produce multiple potential conclusions from a given set of premises. These conclusions could be entirely correct, partially correct, or even contradictory. While our primary focus was on consistency and the presence of heuristics, we also provided an additional analysis of the completeness of the models' answers in the appendix. However, we did not measure model performance in terms of how often the models generated all valid conclusions from a set of premises.

To further analyze the results, adversarial studies could be run to exclude that SFT has learned shortcuts we have not considered here.
We conducted an analysis that is mainly “behavioral” and inspired by the literature on human syllogistic reasoning, however, a more complete analysis should also try to connect these behavioral evidence to the causal mechanisms that drive model behavior.
 

We performed SFT using pseudo-words, and our results suggest that this has an impact in that it helps models reason with form, rather than lexical content, thereby potentially overcoming content biases. A fuller picture could be obtained through experiments with other strategies for disrupting lexical content, such as using random (but real) words. Furthermore, these conclusions would benefit from careful ablation studies. 

\section*{Acknowledgments}
We are grateful to C.J. (Kees) van Deemter, Denis Paperno, Jakub Szymanik, and Roni Katzir for their constructive comments and stimulating questions. We thank Le Dieu Thu and Marco Baroni for their feedback on an earlier version of this work. The work has been supported by a donation by Amazon Alexa to the last author.

\bibliography{raffa}
\appendix
\section{Dataset}
\label{app:data}

\begin{figure*}
\begin{center}
\fbox{%
\begin{tabular}{lll}
\texttt{siameses} & \texttt{cats} & \texttt{felines} \\
\texttt{labradors} & \texttt{dogs} & \texttt{canines} \\
\texttt{anguses} & \texttt{cows} & \texttt{mammals} \\
\texttt{chickadees} & \texttt{birds} & \texttt{winged animals} \\
\texttt{humans} & \texttt{animals} & \texttt{mortals} \\
\texttt{sedans} & \texttt{cars} & \texttt{vehicles} \\
\texttt{cruisers} & \texttt{warships} & \texttt{watercrafts} \\
\texttt{boeings} & \texttt{planes} & \texttt{aircrafts} \\
\texttt{daisies} & \texttt{flowers} & \texttt{plants} \\
\texttt{pines} & \texttt{evergreens} & \texttt{trees} \\
\end{tabular}%
}
\end{center}
\caption{\textbf{Triples of Terms.} Ten triples of terms are used to create believable and unbelievable syllogisms. Each term represents a class of entities and the terms within each triple instantiate a hierarchy of increasing generality, from more specific to broader categories.}\label{fig:triples}
\end{figure*}

\subsection{Valid and Invalid Syllogisms.} Following studies on Human syllogistic reasoning in the psychological literature \citet{khem:theo12, ragni:when19}, out of the 64 syllogistic schemas we consider 27 to be followed by a valid conclusion and the remaining 37 to be followed by no valid conclusions.
The number of syllogisms that are considered valid depends on two assumptions a) we consider all terms to be denoting non-empty sets, and b) we allow for conclusions relating both the terms $a$-$c$ and $c$-$a$. The psychological literature on which we base our analysis also assume both a) and b). Table \ref{tab:all_syllo} shows the valid and invalid syllogistic schemas with their conclusions.

\begin{table*}
\begin{center}
\begin{small}
\begin{tabular}{cllcclcc}
\toprule
\multicolumn{4}{c}{\textbf{Valid Syllogisms}} & \multicolumn{4}{c}{\textbf{Invalid Syllogisms}} \\
\cmidrule(lr){1-4} \cmidrule(lr){5-8}
Syllogism & Premises & Conclusions & Accuracy & Syllogism & Premises & Conclusions & Accuracy \\
\cmidrule(lr){1-4} \cmidrule(lr){5-8}
AA1 & Aab, Abc & Aac, Iac, Ica & 88 & AA3 & Aab, Acb & NVC & 31 \\
AA2 & Aba, Acb & Aca, Iac, Ica & 54 & AI1 & Aab, Ibc & NVC & 16 \\
AA4 & Aba, Abc & Iac, Ica & 16 & AI3 & Aab, Icb & NVC & 37 \\
AI2 & Aba, Icb & Iac, Ica & 90 & AO1 & Aab, Obc & NVC & 14 \\
AI4 & Aba, Ibc & Iac, Ica & 83 & AO2 & Aba, Ocb & NVC & 17 \\
AE1 & Aab, Ebc & Eac, Eca, Oac, Oca & 87 & IA2 & Iba, Acb & NVC & 12 \\
AE2 & Aba, Ecb & Oac & 1 & IA3 & Iab, Acb & NVC & 28 \\
AE3 & Aab, Ecb & Eac, Eca, Oac, Oca & 81 & II1 & Iab, Ibc & NVC & 33 \\
AE4 & Aba, Ebc & Oac & 8 & II2 & Iba, Icb & NVC & 30 \\
AO3 & Aab, Ocb & Oca & 40 & II3 & Iab, Icb & NVC & 51 \\
AO4 & Aba, Obc & Oac & 54 & II4 & Iba, Ibc & NVC & 61 \\
IA1 & Iab, Abc & Iac, Ica & 88 & IO1 & Iab, Obc & NVC & 33 \\
IA4 & Iba, Abc & Iac, Ica & 81 & IO2 & Iba, Ocb & NVC & 49 \\
IE1 & Iab, Ebc & Oac & 44 & IO3 & Iab, Ocb & NVC & 53 \\
IE2 & Iba, Ecb & Oac & 13 & IO4 & Iba, Obc & NVC & 54 \\
IE3 & Iab, Ecb & Oac & 20 & EE1 & Eab, Ebc & NVC & 44 \\
IE4 & Iba, Ebc & Oac & 28 & EE2 & Eba, Ecb & NVC & 44 \\
EA1 & Eab, Abc & Oca & 3 & EE3 & Eab, Ecb & NVC & 76 \\
EA2 & Eba, Acb & Eac, Eca, Oac, Oca & 78 & EE4 & Eba, Ebc & NVC & 66 \\
EA3 & Eab, Acb & Eac, Eca, Oac, Oca & 80 & EO1 & Eab, Obc & NVC & 28 \\
EA4 & Eba, Abc & Oca & 9 & EO2 & Eba, Ocb & NVC & 47 \\
EI1 & Eab, Ibc & Oca & 8 & EO3 & Eab, Ocb & NVC & 49 \\
EI2 & Eba, Icb & Oca & 37 & EO4 & Eba, Obc & NVC & 57 \\
EI3 & Eab, Icb & Oca & 21 & OA1 & Oab, Abc & NVC & 20 \\
EI4 & Eba, Ibc & Oca & 15 & OA2 & Oba, Acb & NVC & 13 \\
OA3 & Oab, Acb & Oac & 36 & OI1 & Oab, Ibc & NVC & 36 \\
OA4 & Oba, Abc & Oca & 42 & OI2 & Oba, Icb & NVC & 31 \\
& & & & OI3 & Oab, Icb & NVC & 49 \\
& & & & OI4 & Oba, Ibc & NVC & 47 \\
& & & & OE1 & Oab, Ebc & NVC & 37 \\
& & & & OE2 & Oba, Ecb & NVC & 51 \\
& & & & OE3 & Oab, Ecb & NVC & 47 \\
& & & & OE4 & Oba, Ebc & NVC & 49 \\
& & & & OO1 & Oab, Obc & NVC & 37 \\
& & & & OO2 & Oba, Ocb & NVC & 42 \\
& & & & OO3 & Oab, Ocb & NVC & 64 \\
& & & & OO4 & Oba, Obc & NVC & 66 \\\bottomrule
\end{tabular}
\end{small}
\end{center}
\caption{\textbf{Valid and Invalid syllogistic schemas with their conclusions.} NVC stands for no valid conclusion. The table is divided into valid syllogisms (27) where there is at least one valid conclusion and invalid syllogisms (37) where there are no valid conclusions. The table is adapted from \citet{khem:theo12}. The accuracy scores on each schema are also based on \citet{khem:theo12}}\label{tab:all_syllo}
\end{table*}

\subsection{Believable and Unbelievable sets.} Figure \ref{fig:triples} shows the 10 triples of terms we used to create the sets of \textit{believable} and \textit{unbelievable} syllogisms. Each of these terms represents a class and the terms in each triple are always in a taxonomic or hierarchical relationship, specifically a "is-a" or "kind-of" relationship. In each row, the first term is a specific instance or subtype, the second term is a more general category or supertype, and the third term is an even more general category or supertype.

To generate syllogisms that are true (\textit{believable}) or false (\textit{unbelievable}) in the real world we always used these 30 terms and re-arranged them accordingly. Triples are re-arranged either by changing the order of terms or by swapping a term from one triple with one belonging to another triple.

Now we will illustrate an example of how we re-arranged a triple to obtain a \textit{believable} syllogism. We can use the first triple (\texttt{{\color{teal}siameses}, {\color{teal}cats}, {\color{teal}felines}}) to create a syllogism of the schema AA1 for the believable set of syllogisms. By substituting the terms $a$, $b$, and $c$ with this triple in the schema AA1 we obtained the premises \texttt{{\color{teal}All siameses are cats}} and \texttt{{\color{teal}All cats are felines}}. These two premises are then followed by the three correct conclusions \texttt{{\color{teal}All siameses are feline, Some siameses are felines, Some felines are siameses}}. All these conclusions are true in the actual world. 

Instead, when dealing with different schemas, we could not use the same triple in the same order and obtain a conclusion that is true in the actual world. For example, for the schema EA3 substituting the terms $a$, $b$, and $c$, with the triple \texttt{{\color{teal}siameses}, {\color{teal}cats}, {\color{teal}felines}} would result in the conclusions \texttt{{\color{teal}No siamese are felines, no felines are siameses, some siameses are not felines, some felines are not siameses.}}, which are all false in the real world. To obtain a syllogism in the schema EA3 which is true in the real world starting from the triple \texttt{{\color{teal}siameses}, {\color{teal}cats}, {\color{teal}felines}} we need to swap \texttt{{\color{teal} siameses}} with a term from another triple, e.g., \texttt{{\color{teal} dogs}}, and swap the order of the terms \texttt{{\color{teal} cats}} and \texttt{{\color{teal} felines}}, obtaining the new modified triple \texttt{{\color{teal}dogs}, {\color{teal}felines}, {\color{teal}cats}}. With this modified triple, we can now substitute it with the terms $a$, $b$, and $c$ in the schema EA3 and obtain the premises \texttt{{\color{teal}No dogs are felines}} and \texttt{{\color{teal}All cats are felines}}. These premises are followed by the conclusions \texttt{{\color{teal}No dogs are cats, no cats are dogs, some dogs are not cats, some cats are not dogs.}} which are all true in the actual world.

We manually modified triples in this way to obtain the sets of believable and unbelievable syllogisms.

\subsection{Syllogisms with 3 and 4 premises.} 

To build datasets of syllogisms with more than two premises we exploited the property of transitivity present in chains of $A$ statements. In fact, given two terms $a$ and $b$ plus $c_1, ..., c_n$ auxiliary terms, we can construct arbitrarily long sequences of $A$ premises from which we can derive $Aab$: $Aac_1,Ac_1c_2, ... , Ac_{n-1}c_n, Ac_nb$ (for $n \geq 1$).

We built datasets with 3 and 4 premises by substituting A premises of schemas that contain at least one A premise with sequences of 2 and 3 A premises, respectively. Since these datasets are built to test models on longer inferences and not to check whether they suffer from biases derived from the semantic content on terms present in the syllogisms, we used pseudowords with no real meaning.

The following is an example of how a schema can be extended from the initial 2 premises to 3 and 4 premises. Let's consider a syllogism from the schema AE1 with pseudoword terms, \texttt{{\color{teal}All schluiev are cycleewn}} as first premise, and \texttt{{\color{teal}No cycleewn are tsaongly}} as second premise. If we want to obtain a syllogism with 3 premises in total and with the same conclusion as AE1, we can substitute \texttt{{\color{teal}All schluiev are cycleewn}} with \texttt{{\color{teal}All schluiev are gleows, All gleows are cycleewn}}. From these 2 premises we can derive back \texttt{{\color{teal}All schluiev are cycleewn}}. Instead, If we want to obtain a syllogism with 4 premises in total and the same conclusion as AE1, we can substitute \texttt{{\color{teal}All schluiev are cycleewn}} with \texttt{{\color{teal}All schluiev are gleows, All gleows are voost, All voost are cycleewn}}. From these 3 premises we can derive back \texttt{{\color{teal}All schluiev are cycleewn}}.

\section{Implementational Details}
\label{app:impl}

\subsection{Computational resources}

All our experiments were carried out using a single server with three 24GB NVIDIA GPUs: 2 RTX A5000 and 1 Quadro P6000. The total computation time to fine-tune models and run all evaluations is approximately 72 hours.

\subsection{Training}

\paragraph{Supervised Fine-tuning.} To fine-tune models we used the popular parameter-efficient method LoRA \citep{hu:lora21}. We set the rank 32
and applied LoRA to the query and key attention matrices. We used no bias and set LoRA’s $\alpha$ to 16 and dropout to 0.05. To select the rank
we compared the performance of different ranks (8, 16, 32) and found 32 to be the best value. We used
AdamW as optimizer with a learning rate of 5e-4. The other optimizer’s hyper-parameters are left in their default setting. We used a batch 
size of 2 with 32 gradient accumulation steps. Fine-tunings were performed using \texttt{PyTorch} and LoRA layers were created and added 
to the original models using the built-in methods from the libraries \texttt{transformers} and \texttt{peft}.

As fine-tuning objective, we minimize the cross-entropy loss on text sequences composed by the two premises, the list of options presented in random order, and the correct conclusions. At test time, fine-tuned models are prompted with the two premises plus the list of options in random order and have to complete the sequence generating conclusions.

Models are fine-tuned for 100 epochs and evaluate models' performance at the end of each epoch using a small development set. This development set is built using the 64 schemas and pseudo-words that are unseen during training. The SFT models used in the main paper are the ones that obtained the best accuracy on this development set.







\begin{figure*}[t]
\begin{center}
\begin{tikzpicture}
\node[draw, rounded corners, text width=.8\linewidth, align=flush left, inner sep=12 pt] (box) {
    You will be presented with premises together with eight possible conclusions and the option 'Nothing follows'.
  
    Write the conclusion that logically follows given the premises or 'Nothing follows' 
    
    if none of the other conclusions logically follow.
    
    Read the passage of information thoroughly and select the correct answer from the available options.
    
    Read the premises thoroughly to ensure you know what the premise entails.
};
\end{tikzpicture}
\end{center}
\caption{\textbf{Task instruction prompt.} We adapted our prompt from \citet{eisape:syst23} and added the additional strings \textit{“Read the passage of information thoroughly and select the correct answer from the available options. Read the premises thoroughly to ensure you know what the premise entails.”} to make the task requirements more explicit.} \label{fig:prompt}
\end{figure*}

\subsection{Inference} 

\paragraph{Text Generation Details.} For all test sets and settings, we use greedy decoding to generate syllogism conclusions from LMs. We let LMs generate 20 tokens for each answer, which can correspond to 2 or 3 conclusions based on the terms appearing in the conclusions. Since it is possible to derive from 1 to 4 possible conclusions depending on which syllogism is considered (e.g. AE1 has four conclusions, OA1 has one, see Table \ref{tab:all_syllo}), we set the number of generated tokens to an amount that can let models generate more than one conclusion. This choice was motivated by the fact that considering only most probable conclusion would limit a deeper understanding of how LMs complete the task. In fact, generating only one conclusion would prevent us from analyzing whether models contradict themselves (Table \ref{tab:unbe-cont} center) or whether they follow the Atmosphere Theory (Table \ref{tab:unbe-cont} right). For the ZS-CoT setting, we let pretrained LMs generate a reasoning chain of 50 tokens and instruction-tuned LMs a chain of 70 tokens. This was done since we observed that instruction-tuned models tend to be more verbose and need on average more tokens to complete their reasoning. For experiments where syllogisms with 3 and 4 premises were involved, we let all models generate 70 tokens.

\paragraph{Answer extraction.} Since we frame the task as multiple-choice with LMs generating conclusions, we tried to constrain the generation to make models output only the desired answer and nothing else. For both the ICL settings, we added at the end of the prompt an additional string to elicit a direct answer by models. The additional prompt is \textit{“Given the premises I choose the following option(s):”}. Instead, for the ZS-CoT setting we first conditioned the model to output a reasoning chain by adding the string \textit{“Let's think this through, step by step”} and then we extracted a final answer by adding the string \textit{“So, my final answer(s) is/are:"} at the end of the reasoning chain. SFT models needed no additional prompts since they were fine-tuned on the task.

\paragraph{Weigths Precision.} Due to the limited computational resources we used 8-bit quantization for LLaMA-2-13B and 4-quantization for LLaMA-3-70B, for both pretrained and instruction-tuned models. For all other models, we used half-precision inference.

\subsection{Prompts}

Figure \ref{fig:prompt} shows the task instruction prompt used for non-fine-tuned models (ZS-CoT, ICL$_{in}$, ICL$_{out}$). Instead, for models directly fine-tuned on the multiple-choice task (FT) no instruction prompt is used. We adapted our prompt from \citet{eisape:syst23} and added the additional strings \textit{“Read the passage of information thoroughly and select the correct answer from the available options. Read the premises thoroughly to ensure you know what the premise entails.”} to make the task requirements more explicit.

In addition to the task instruction prompt, the prompts in the ICL settings provide in-context examples to models. Figure \ref{fig:out_prompt} and Figure \ref{fig:in_prompt} show an example of these prompts for ICL$_{out}$ and ICL$_{in}$, respectively. These in-context examples are in both settings built using pseudo-words, whereas they differ in how they relate to the test example: in ICL$_{out}$ they are generated randomly making sure that they do not instantiate the schema in the test example; for ICL$_{in}$ instead, they share the same schema with the test example.

The number of in-context examples in both settings is set to 5. This number of examples was chosen to provide enough diverse (ICL$_{out}$) or similar (ICL$_{in}$) examples to models while limiting the amount of computation required to process a larger input.

\cut{\begin{table*}
\begin{tabular}{cc}
\begin{tabular}{c|c|cc|c|c}
  & Humans & ML8 & ML1 & RAG & Chance\\\hline
AI2 & 90 & 100 & 30 & 100 & 20 \\
AA1 & 88 & 100 & 100 & 90 & 38 \\
IA1 & 88 & 100 & 60 & 90 & 20 \\
AE1 & 87 & 100 & 70 & 90 & 50 \\
AI4 & 83 & 100 & 60 & 100 & 20 \\
IA4 & 81 & 100 & 100 & 80 & 20 \\
AE3 & 81 & 90 & 60 & 90  & 50 \\
EA3 & 80 & 90 & 100 & 80 & 50 \\
\end{tabular}
&
\begin{tabular}{c|c|cc|c|c}
  & Humans & ML8 & ML1 & RAG & Chance\\\hline
AE2 & 1 & 100 & 80 & 40 & 13 \\
EA1 & 3 & 100 & 90 & 20 & 13 \\
AE4 & 8  & 100 & 90 & 40 & 13 \\
EI1  & 8 & 100 & 40 & 50 & 13 \\
EA4 & 9 & 100 & 80 & 90 & 13 \\
IE2 & 13 & 100 & 90 & 90 & 13 \\
EI4 & 15 & 100 & 60 & 80 & 13 \\
AA2 & 16 & 100 & 40 & 90 & 38 \\
\end{tabular}
\end{tabular} \caption{\rb{OLD NUMBERS}The valid syllogism schemas: easier (LEFT),
  and harder (RIGHT) for humans  (based
  on the meta-review by~\citet{khem:theo12}).} \label{tab:easysyl}
\end{table*}
}

\begin{figure*}[t]
\begin{center}
\begin{tikzpicture}
\node[draw, rounded corners, text width=.9\linewidth, align=flush left, inner sep=16 pt] (box) {
\small
\sethlcolor{mistyrose}\hl{\textbf{INSTRUCTION}}

You will be presented with premises together with eight possible conclusions and the option 'Nothing follows'. Write the conclusion that logically follows given the premises or 'Nothing follows' if none of the other conclusions logically follow. Read the passage of information thoroughly and select the correct answer from the available options. Read the premises thoroughly to ensure you know what the premise entails.
\vspace{4pt}

\sethlcolor{seafoam}
\hl{\textbf{CONTEXT}}

Use the examples in the following context to better answer the test examples:

\vspace{4pt}
Syllogism:

Premise 1: Some qeabs are mcclaiands.
Premise 2: Some khoists are not mcclaiands.

<shuffled-options>

Answer: Nothing follows.

\vspace{4pt}
Syllogism:

Premise 1: Some ghuists are not cziaords.
Premise 2: No synaarg are cziaords.

<shuffled-options>

Answer: Nothing follows.

\vspace{4pt}
Syllogism:

Premise 1: All klaabs are frugh.
Premise 2: Some klaabs are khusch.

<shuffled-options>

Answer: Some khusch are frugh or some frugh are khusch.

\vspace{4pt}
Syllogism:

Premise 1: Some schwiong are not wuolls.
Premise 2: Some wuolls are psycheift.

<shuffled-options>

Answer: Nothing follows.

\vspace{4pt}
Syllogism:

Premise 1: Some sneitt are not mcduiends.
Premise 2: All sneitt are schriant.

<shuffled-options>

Answer: Some schriant are not mcduiends.

\vspace{4pt}
\sethlcolor{warmcream}
\hl{\textbf{TEST}}

Complete the following test example:

\vspace{4pt}
Syllogism:

Premise 1: All siameses are cats.
Premise 2: All cats are felines.

Options:
All felines are siameses.
Nothing follows.
Some siameses are not felines.
Some felines are not siameses.
No felines are siameses.
All siameses are felines.
Some siameses are felines.
No siameses are felines.
Some felines are siameses.

Answer:
};
\end{tikzpicture}
\end{center}
\caption{\textbf{ICL$_{out}$ Prompt.} Example of a prompt used in the ICL$_{out}$ setting. First, a \sethlcolor{mistyrose} \hl{task instruction} is presented to the model. The task instruction is then followed by 5 randomly chosen \sethlcolor{seafoam} \hl{in-context demonstrations} where pseudo-words are used in place of variables. These demonstrations have a different schema compared with the one in the test syllogism. Finally, the \sethlcolor{warmcream} \hl{test example} is presented to the model which starts generating an answer after the string “Answer:". In the above example, the nine options for each in-context example are substituted with the special token \texttt{<shuffled-options>} for ease of presentation.}\label{fig:out_prompt}
\end{figure*}

\begin{figure*}[t]
\begin{center}
\begin{tikzpicture}
\node[draw, rounded corners, text width=.9\linewidth, align=flush left, inner sep=16 pt] (box) {
\small
\sethlcolor{mistyrose}\hl{\textbf{INSTRUCTION}}

You will be presented with premises together with eight possible conclusions and the option 'Nothing follows'. Write the conclusion that logically follows given the premises or 'Nothing follows' if none of the other conclusions logically follow. Read the passage of information thoroughly and select the correct answer from the available options. Read the premises thoroughly to ensure you know what the premise entails.
\vspace{4pt}

\sethlcolor{seafoam}
\hl{\textbf{CONTEXT}}

Use the examples in the following context to better answer the test examples:

\vspace{4pt}
Syllogism:

Premise 1: All shucts are gloogy.
Premise 2: All gloogy are kriurs.

<shuffled-options>

Answer: All shucts are kriurs or some shucts are kriurs or some kriurs are shucts.

\vspace{4pt}
Syllogism:

Premise 1: All diks are kheaunk.
Premise 2: All kheaunk are chrusk.

<shuffled-options>

Answer: All diks are chrusk or some chrusk are diks or some diks are chrusk.

\vspace{4pt}
Syllogism:

Premise 1: All syniaock are mieg.
Premise 2: All mieg are mcceaubs.

<shuffled-options>

Answer: All syniaock are mcceaubs or some syniaock are mcceaubs or some mcceaubs are syniaock.

\vspace{4pt}
Syllogism:

Premise 1: All greedy are wuays.
Premise 2: All wuays are gaix.

<shuffled-options>

Answer: All greedy are gaix or some greedy are gaix or some gaix are greedy.

\vspace{4pt}
Syllogism:

Premise 1: All bliodly are sqauecks.
Premise 2: All sqauecks are raills.

<shuffled-options>

Answer: All bliodly are raills or some bliodly are raills or some raills are bliodly.

\vspace{4pt}
\sethlcolor{warmcream}
\hl{\textbf{TEST}}

Complete the following test example:

\vspace{4pt}
Syllogism:

Premise 1: All siameses are cats.
Premise 2: All cats are felines.

Options:
All felines are siameses.
Nothing follows.
Some siameses are not felines.
Some felines are not siameses.
No felines are siameses.
All siameses are felines.
Some siameses are felines.
No siameses are felines.
Some felines are siameses.

Answer:
};
\end{tikzpicture}
\end{center}
\caption{\textbf{ICL$_{in}$ Prompt.} Example of a prompt used in the ICL$_{in}$. First, a \sethlcolor{mistyrose} \hl{task instruction} is presented to the model. The task instruction is then followed by 5 \sethlcolor{seafoam} \hl{in-context demonstrations} with the same schema as in the test example. Finally, the \sethlcolor{warmcream} \hl{test example} is presented to the model which starts generating an answer after the string “Answer:". In the above example, the nine options for each in-context example are substituted with the special token \texttt{<shuffled-options>} for ease of presentation}\label{fig:in_prompt}
\end{figure*}

\section{Experimental Details}
\label{app:exp}

\paragraph{Task Accuracy} The accuracies shown in Tables \ref{tab:zeroshot}, \ref{tab:learning}, and \ref{tab:length} are computed by allowing models to generate conclusions and considering an answer successful when it contains at least one of the correct conclusions. As the in-depth analysis of model consistency and the overlap with the Atmosphere heuristic show (see Table \ref{tab:unbe-cont}), Pythia and LLaMA in their different settings also generate incorrect answers. For this reason, we also show the Top 1 accuracy obtained by the models, computed by retaining only the first generated answer as the model prediction. The results are shown in Table \ref{tab:top1} and are computed by selecting the best run for each setting out of the three reported in Table \ref{tab:learning}.

\begin{table}
\begin{center}
    \small
    \begin{tabular}{clcccc}
    \toprule
    &  & \multicolumn{3}{c}{\textbf{Believable}} & \textbf{Unbelievable} \\
    \cmidrule(lr){3-5} \cmidrule(lr){6-6}
    & & Acc & Valid & Invalid & Valid \\\midrule
    \multirow{4}{*}{\rotatebox{90}{\sc Pythia}} & ZS-CoT & 7.03 & 16.30 & 0.27 & 18.89 \\
    & ICL$_{in}$ & 13.44 & 27.41 & 3.24 & 26.67 \\
    & ICL$_{out}$ & 7.81 & 17.41 & 0.81 & 15.93 \\
    & SFT & 78.59 & 62.59 & 90.27 & 63.33 \\
    \midrule
    \multirow{4}{*}{\rotatebox{90}{\sc LLaMA}} & ZS-CoT & 27.34 & 63.70 & 0.81 & 57.04 \\
    & ICL$_{in}$ & 31.41 & 72.96 & 1.08 & 70.74 \\
    & ICL$_{out}$ & 24.69 & 58.15 & 0.27 & 50.74 \\
    & SFT & 91.88 & 81.11 & 99.73 & 90.00 \\
    \bottomrule
        \end{tabular}
      \caption{\textbf{Top 1 Accuracy}. We compute the Top 1 Accuracy by retaining only the first generated answer as the prediction. The results are from the best run for each setting out of the three reported in the main paper.}\label{tab:top1}
    \end{center}
\end{table}

\paragraph{Correlation between humans and models} Tables \ref{tab:zeroshot} and \ref{tab:learning} show the correlation between the accuracies obtained by humans and models on valid syllogisms of the \textit{believable} set. The human accuracies we used for each schema are taken from \citet{khem:theo12} and can be seen in Table \ref{tab:all_syllo}. Regarding the accuracies of LMs, the \textit{believable} set has 10 instances of the same schema with different terms. Given this property of the set, it is straightforward to compute the accuracy of schemas appearing in them. After having obtained an accuracy for each valid schema, we then compute the Spearman correlation coefficient between humans and models.

A higher correlation means that models and humans struggle with the same schemas. This can be true even when models have on average a higher accuracy compared with humans. For example, LLaMA-3 70B has an accuracy of 58.89\% on valid syllogisms whereas humans have an accuracy of 44.63\%, but the correlation between the two is 0.80. This situation occurs if both models and humans find the same syllogisms \textit{relatively} easier or harder, compared to their own average performance. In other words, the syllogisms that humans struggle with the most tend to also be challenging for models, and those that humans find easiest tend to be easier for models too - even though the models get more answers right overall.

\begin{figure*}[t]
\includegraphics[width=0.8\linewidth]{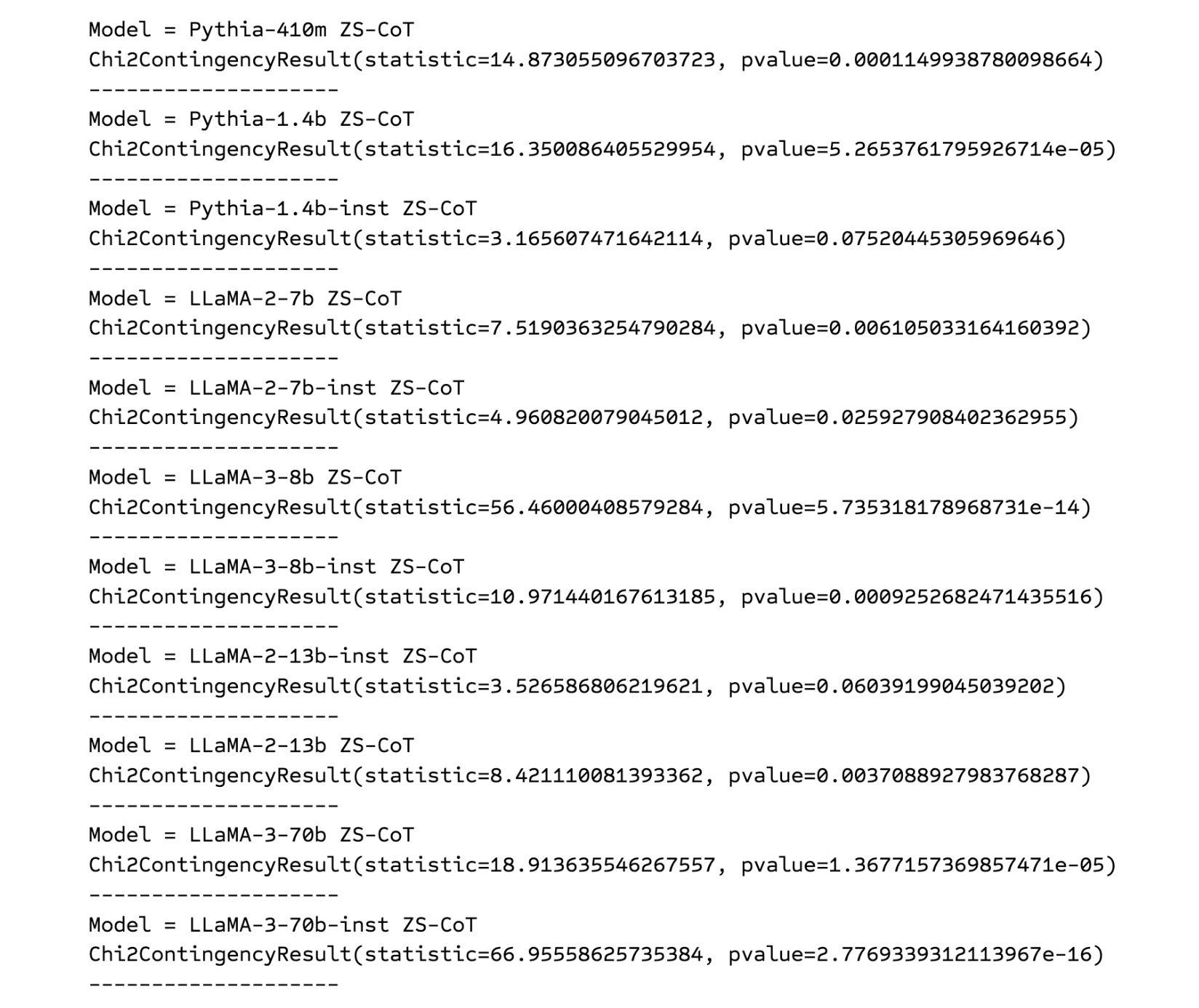}
\centering
\caption{\textbf{Results of $\chi^2$'s Test on Content Effect.} We employed a significance level of 0.05.}\label{fig:stat}
\end{figure*}

\paragraph{Analysis of Content Effect} We report in Figure \ref{fig:stat} the results of the statistical tests used to verify whether the performance differences for the content effect from Table \ref{tab:zeroshot} are significant. We employed the \href{https://docs.scipy.org/doc/scipy/reference/generated/scipy.stats.chi2_contingency.html#rf346382074c5-2}{\texttt{scipy}} implementation of the Chi-square test for independence of variables in a contingency table. This test uses Pearson's chi-squared statistic with Yates' continuity correction. 

In addition to this statistical analysis, we show here the results of an additional experiment measuring the content effect in ZS-CoT, ICL$_{in}$, ICL$_{out}$, and SFT settings. A model suffering from content effect bias would prefer a believable conclusion over an unbelievable one. Hence, we check how often models generate believable conclusions when the correct one is unbelievable ($B | U$), and, conversely, how often models generate unbelievable conclusions when the correct one is believable ($U | B$). A model that suffers from content bias would produce $B | U > U | B$. Table \ref{tab:content_annotation} shows
that all settings but SFT produce more $B | U$, strengthening the results of Table 2.

In order to compute $U | B$ and $B | U$ for each setting, we used LLaMA-3-inst 8B to annotate the generated conclusions. To verify the reliability of the model we manually annotated 441 conclusions with LLaMA-3-inst 8B obtaining 93.42\% accuracy w.r.t human judgment.

\begin{table}
    \centering
\begin{tabular}{clcc}
\toprule
\multicolumn{2}{l}{} & \multicolumn{2}{c}{\textbf{Content Effect}} \\
 \cmidrule(lr){3-4}
 \multicolumn{2}{l}{} &  $U | B$  & $B | U$ \\\midrule
\multirow{4}{*}{\rotatebox{90}{\sc Pythia}} & ZS-CoT & 35.31 & 52.45 \\ 
& ICL$_{out}$ & 35.64 & 50.77 \\ 
& ICL$_{in}$ & 34.64 & 48.44 \\
& SFT & 19.81 & 21.39 \\ \midrule 
\multirow{4}{*}{\rotatebox{90}{\sc LLaMA}} & ZS-CoT & 11.66 & 22.10 \\
& ICL$_{out}$ & 20.95 & 38.94 \\
& ICL$_{in}$ & 11.59 & 22.94 \\
& SFT & 1.06 & 4.28 \\\bottomrule 
\end{tabular}
\caption{\textbf{Content Effect bias} (Left): We compute how often the model generates an unbelievable option while the ground truth conclusion is believable ($U|B$), and vice-versa ($B|U$). A higher percentage in the latter case signals a content effect bias.}\label{tab:content_annotation}
\end{table}

\section{Inconsistency vs. Incompleteness}
\label{app:inc}

In this additional analysis, we explore a complementary aspect of the consistency of generated conclusions: completeness. While the main paper focuses on consistency, as we consider it a more fundamental property for testing syllogistic inference in LLMs, here we examine how well the models generate all relevant conclusions. Our reasoning is that even if a logical agent cannot produce every correct conclusion from a set of premises, it should at least ensure the conclusions it does generate are not contradictory.

As noted in the limitations section, we did not require the models to generate every possible conclusion from a given set of premises and then assess accuracy based on whether they correctly listed all and only valid conclusions (a stricter measure of performance). Instead, to gain insight into the completeness of the generated answers, we measured how often a model produced both a conclusion and its converse when the relationship between the terms was symmetrical. In syllogistic logic, there are two such conclusions: I (Some \textit{a} are \textit{b}) and E (No \textit{a} are \textit{b}).

\begin{table}
\begin{center}
\small
\begin{tabular}{clccc}
\toprule
&  & \multicolumn{3}{c}{\textbf{Completeness}} \\
\cmidrule(lr){3-5}
\multicolumn{2}{l}{} & \% Inc & \% Inc(I) & \% Inc(E) \\\midrule
\multirow{4}{*}{\rotatebox{90}{\sc Pythia}} & ZS-CoT & 83.46 & 78.62 & 90.45 \\
& ICL${out}$ & 83.52 & 82.34 & 84.85 \\
& ICL${in}$ & 81.20 & 78.69 & 83.67 \\
& SFT & 90.46 & 99.32 & 82.79 \\ \midrule 
\multirow{4}{*}{\rotatebox{90}{\sc LLaMA}} & ZS-CoT & 96.93 & 99.05 & 95.51 \\
& ICL${out}$ & 72.93 & 96.08 & 69.54 \\
& ICL${in}$ & 76.24 & 83.52 & 74.25 \\
& SFT & 43.14 & 61.36 & 29.31 \\ \bottomrule
\end{tabular}
\caption{\textbf{Completeness}. Performance comparison of LLaMA and Pythia showing overall percentages of incomplete answers, as well as the percentage of incomplete answers for I and E conclusions. In this table lower is better.}\label{tab:incompleteness}
\end{center}
\end{table}

Table \ref{tab:incompleteness} shows the proportion of incomplete answers across all responses provided by LLaMA-3 8B and Pythia 1.4B on the believable set. The table also breaks down the ratio of incomplete answers by conclusion type (I and E). Unlike consistency (see Table \ref{tab:unbe-cont}), where both models performed better, they generated significantly more incomplete answers, with LLaMA generally outperforming Pythia, except in the ZS-COT setting. Similar to consistency results, the two fine-tuned models showed differing patterns: Pythia produced fewer complete answers (90.46\% vs. 83.46\%, 83.45\%, 81.20\%), while LLaMA provided more complete responses (43.14\% vs. 96.93\%, 72.93\%, 76.24\%). Interestingly, LLaMA’s ICL setting improved completeness (72.93\%, 76.24\% vs. 96.93\%) while slightly reducing consistency (see Table \ref{tab:unbe-cont}).

\begin{figure*}[t]
\includegraphics[width=0.7\linewidth]{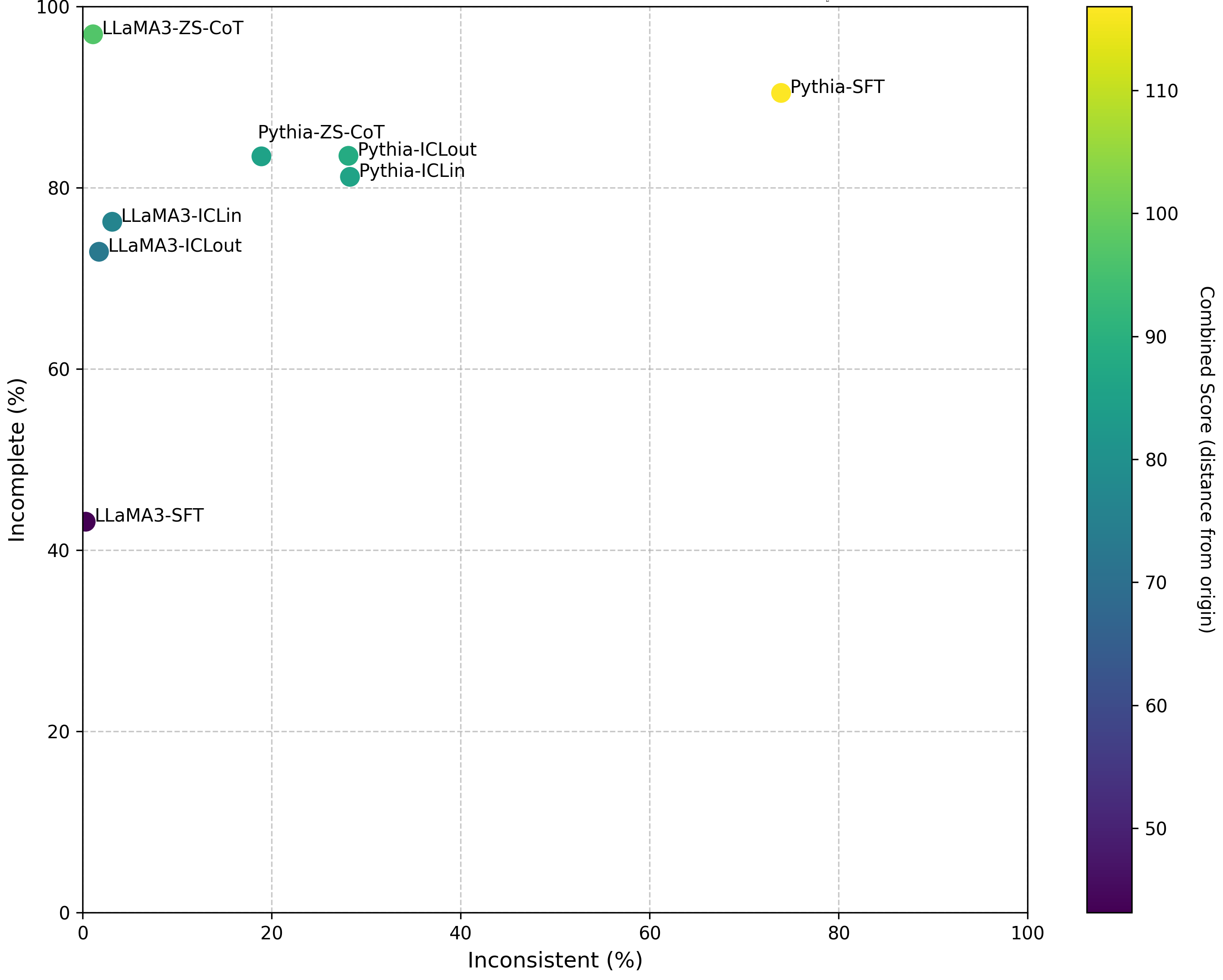}
\centering
\caption{\textbf{Incompleteness vs. Inconsistency}. Plot of inconsistent and incomplete answers provided by the LLaMA-3 8B and Pythia 1.4B in all settings. The x-axis measures the percentage of inconsistent answers, while the y-axis measures the percentage of incomplete answers. A color scale on the right indicate the distance from the origin, which represents a model that is both complete and consistent.}\label{fig:vs_plot}
\end{figure*}

Figure \ref{fig:vs_plot} presents a 2-D visualization of the percentages of inconsistent and incomplete answers provided by both models across all settings. The x-axis shows the percentage of inconsistent answers, while the y-axis represents the percentage of incomplete answers. Additionally, a color scale on the right indicates the distance from the origin, representing a complete and consistent model. This plot highlights the differences between LLaMA-3 8B and Pythia 1.4B models. For LLaMA-3 8B, the model is highly consistent in all settings, and most of the variance between settings is explained by increased answer completeness, which is positively correlated with accuracy. The settings are ranked in the same way (SFT > ICL > CoT) for both raw accuracy and completeness. In contrast, the variance between settings for Pythia 1.4B is primarily explained by inconsistency, which is negatively correlated with accuracy. In terms of raw accuracy, the model settings are ranked SFT > ICL > CoT, whereas for consistency, the ranking is CoT > ICL > SFT.

\section{Heuristics Theories of Syllogistic Reasoning}
\label{app:heur}

Among the theories of syllogistic reasoning proposed in the cognitive psychology literature, we report the proportion of mistakes predicted by the Atmosphere theory of syllogistic reasoning for valid and invalid syllogisms. The Atmosphere theory is among the theories of syllogistic reasoning proposed in the cognitive science literature that is based on heuristics. In addition to the proportion of mistakes predicted by the Atmosphere theory, we show here also the proportion of mistakes predicted by other popular theories of syllogistic reasoning based on heuristics and reviewed in \citet{khem:theo12}.

\paragraph{Atmosphere.} The Atmosphere theory is derived from the hypothesis that reasoners are biased toward conclusions that fit the mood of the premises. This hypothesis was first proposed by \citet{wood:1935} to account for human syllogistic reasoning. \citet{revlis:1975} created a rule-based feature selection model that follows the Atmosphere hypothesis. The model's deductions are based on a feature-matching scheme that requires no understanding of the meaning of the two premises.

To determine which conclusions follow from two premises the model first extracts from the premises two binary features called \textit{quantity} and \textit{polarity}. Quantity is positive $(+)$ if the premise is universal and negative $(-)$ otherwise. Instead, polarity is positive $(+)$ if the premise is affirmative and negative $(-)$ otherwise. For instance, each syllogism with two AA premises is represented as $(+,+)$\&$(+,+)$, whereas each syllogism with two EO premises is represented as $(+,-)$\&$(-,-)$. Consequently, the representation of the conclusions is derived using two rules. \textit{Rule 1}: If the two premises have the same sign on a feature, the same sign is also derived for that feature in the conclusion; \textit{Rule 2}: If there are two different signs for a feature, the derived sign for that feature in the conclusion is minus. If we apply the rule to the previously introduced AA and EO pairs of premises, we can derive $(+,+)$ from AA and $(-,-)$ from EO. Since the theory makes no distinction in the order of the \textit{a} and \textit{c} terms in the conclusion, for every AA schema we can derive the conclusions Aac and Aca, whereas for every EO schema, we can derive the conclusions Oac and Oca (See the prediction of the Atmosphere theory for every schema in Table \ref{tab:atmosphere_pred})  

The main limit of the Atmosphere theory is that according to this heuristic, there are no syllogisms that are invalid, i.e. where no valid conclusion follows. This limits the explanatory power of the theory for human syllogistic reasoning since human subjects can correctly respond that “nothing follows” for certain invalid schemas. In addition, subjects can also fail to draw a valid conclusion that fits the atmosphere and wrongly respond that “nothing follows”.

\begin{table}
\begin{center}
\small
\begin{tabular}{llc}
\toprule
\multicolumn{3}{c}{\textbf{Statement Representations}}  \\
\midrule
A: & All A are C | All C are A & $(+, +)$ \\
E: & No A are C | No C are A & $(+, -)$ \\
I: & Some A are C | Some C are A & $(-, +)$ \\
O: & Some A are not C | Some C are not A & $(-, -)$ \\
\midrule
\midrule
\multicolumn{2}{c}{\textbf{Premise pair}} & \textbf{Conclusion} \\
\cmidrule(lr){1-2} \cmidrule(lr){3-3}
AA: & \quad \quad \quad $(+, +)$ \& $(+, +)$ & $(+, +)$\\
AE: & \quad \quad \quad $(+, +)$ \& $(+, -)$ & $(+, -)$ \\
AI: & \quad \quad \quad $(+, +)$ \& $(-, +)$ & $(-, +)$ \\
AO: & \quad \quad \quad $(+, +)$ \& $(-, -)$ & $(-, -)$ \\
EI: & \quad \quad \quad $(+, -)$ \& $(-, +)$ & $(-, -)$ \\
EO: & \quad \quad \quad $(+, -)$ \& $(-, -)$ & $(-, -)$ \\
EE: & \quad \quad \quad $(+, -)$ \& $(+, -)$ & $(+, -)$ \\
II: & \quad \quad \quad $(-, +)$ \& $(-, +)$ & $(-, +)$ \\
IO: & \quad \quad \quad $(-, +)$ \& $(-, -)$ & $(-, -)$ \\
OO: & \quad \quad \quad $(-, -)$ \& $(-, -)$ & $(-, -)$ \\
\bottomrule
\end{tabular}
\end{center}
\caption{\textbf{Atmosphere Theory predictions.} Table adapted from \citet{revlis:1975}.}\label{tab:atmosphere_pred}
\end{table}

\paragraph{Other Heuristics Theories of Syllogistic Reasoning.} Other than the Atmosphere Theory, other three theories of syllogistic reasoning based on heuristics are discussed in \citet{khem:theo12}: \textit{Matching}, \textit{Illicit Conversion}, and \textit{Probability Heuristics}. Table \ref{tab:heur_preds} shows the conclusions predicted by each theory.

Similar to the Atmosphere theory, the \textbf{Matching} theory \cite{wethe:1995} is based on the hypothesis that individuals do not reason logically but exploit a pattern-matching strategy. This strategy involves selecting a conclusion that connects the end terms and aligns with the mood of the more conservative premise, which assumes fewer entities. The statement \textit{No A are B} is the most conservative since it assumes no entities, whereas \textit{All A are B} is the least conservative. The statements \textit{Some A are B} and \textit{Some A are not B} are equally conservative and fall between the two extremes. The Matching theory also never predicts that \textit{nothing follows} for any combination of premises.

The \textbf{Illicit Conversion} theory is instead derived from the observation that human subjects often treat A and O statements as symmetric, making invalid conversions from \textit{All X are Y} to \textit{All Y are X} and from \textit{Some X are not Y} to \textit{Some Y are not X}, and vice-versa. A possible explanation for this behavior is that such conversions frequently yield true conclusions in daily life, even though they are not logically valid. This idea was first proposed by \citet{chapman:1959} and then \citet{revlis:1975} translated it into a model of syllogistic reasoning.

The \textbf{Probability Heuristics} theory was proposed by \citet{chater:1999} and is based on a probability heuristics model (PHM) of syllogisms. This model is built on the assumptions that a) the fallacious reasoning of everyday life should find an account in theories and b) formal logic does not explain how deductive tasks are performed in the psychological laboratory. The solution is that the appropriate theory should be not rooted in logic but in probability theory. The model substitutes the notion of logical validity with the one of probabilistic validity (probabilistically valid statements are called \textit{“p-valid”}), which is based on the conditional probability of one end term given the other. In this model, the probability of a conclusion depends on the probabilities of each premise, and the figure of the syllogism. According to \citeauthor{chater:1999}, individuals do not compute p-validity in their mind but exploit simpler heuristics based on the notion of \textit{“p-entailments”}. In fact, these heuristics converge on p-valid conclusions without requiring an explicit computation of p-validity.

\begin{figure*}[t]
\includegraphics[width=0.8\linewidth]{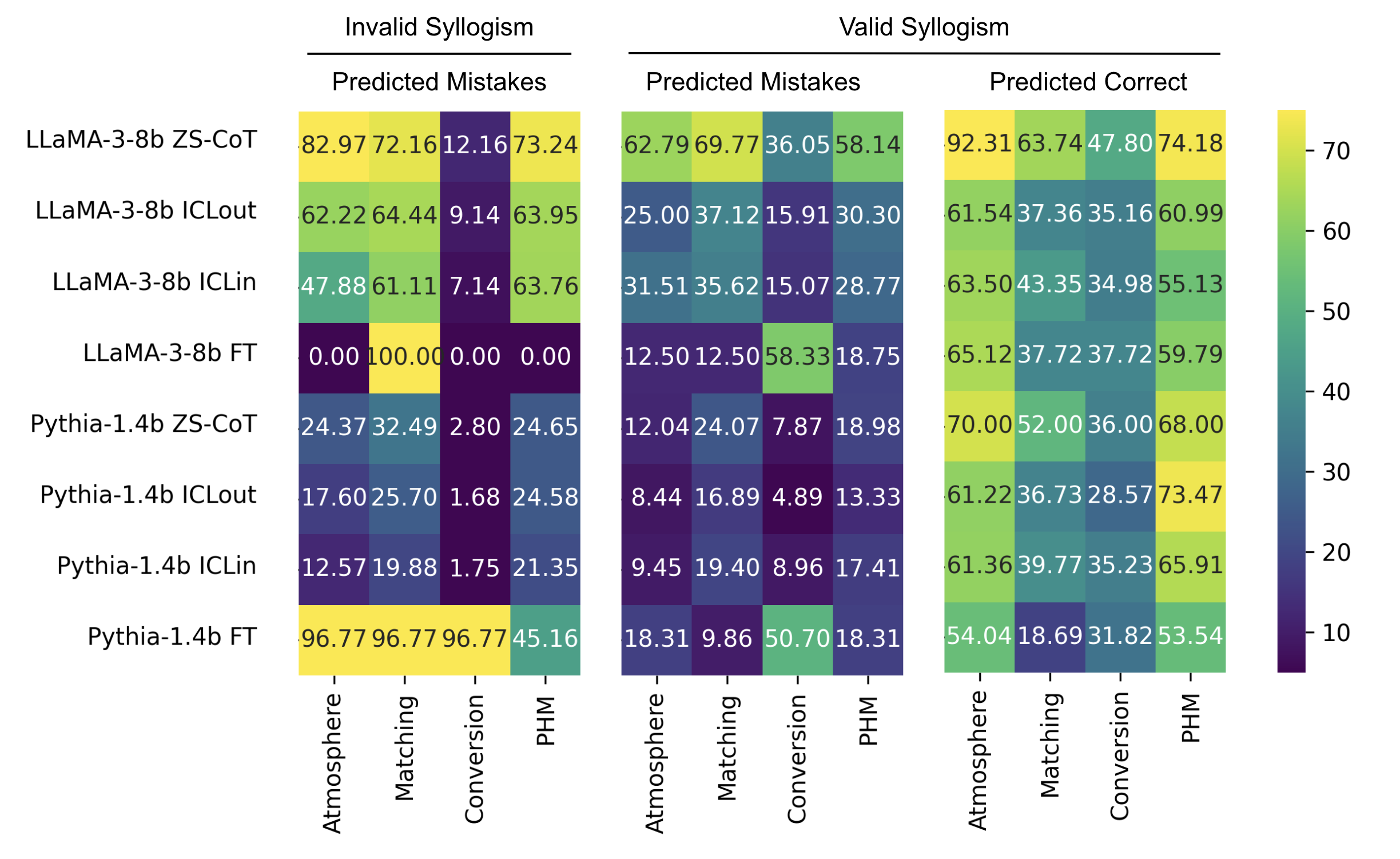}
\centering
\caption{\textbf{Heuristics predictions.} Proportion of mistakes and correct conclusions generated by models that are predicted by heuristic theories.}\label{fig:heur_overlap}
\end{figure*}

\paragraph{Proportion of Answers predicted by heuristic theories.} In Table \ref{tab:atmosphere_pred} (left) we reported the proportion of incorrect conclusions produced by the model that are predicted by the atmosphere theory, for both valid and invalid syllogisms. We show in Figure \ref{fig:heur_overlap} an extension of this analysis to the theories of Illicit Conversion, Matching, and Probability Heuristics. In the main paper, we focused on the Atmosphere theory since it turns out to be the most insightful. In fact, it is the only theory that can explain a high proportion of correct valid conclusions (92\%) as well as mistakes in invalid (82.97\%) and valid (62.79\%) schemas for a model (LLaMA-3 8B ZS-CoT). Although we show the prediction for both correct and incorrect conclusions, it should be noted that the models' error rates (and, conversely, the accuracies) can be rather different, e.g., LLaMA SFT generates wrong conclusions on valid syllogisms in only 48/329 conclusions, while LLaMA ZS-CoT does so in 86/268 conclusions.

Another noticeable pattern is that for both SFT models the Conversion theory can predict 58.33\% (LLaMA-3) and 50.70\% (Pythia) of mistakes on valid syllogisms. However, the theory is not as good at predicting the mistakes of these models on invalid syllogism and their correct conclusions on valid ones. Thus, further analysis would be required to understand whether the models are indeed following the Conversion heuristics or not.

To better contextualize these results, we show in Table \ref{tab:heur_gt} the percentages of ground truth conclusions that are also predicted by these heuristics. A noticeable pattern is that all these heuristics, except for the Conversion theory, fail to capture that “Nothing follows” from invalid syllogisms, whereas on valid syllogisms the best can predict only 62.50\% of correct conclusions (Atmosphere theory).

\begin{table}[h]
\centering
\begin{tabular}{lcc}
\toprule
\textbf{Heuristic} & \textbf{Valid (\%)} & \textbf{Invalid (\%)} \\
\midrule
Atmosphere & 62.50 & 0.00 \\
Matching & 45.83 & 0.00 \\
Conversion & 33.33 & 86.11 \\
PHM & 60.42 & 0.00 \\
\bottomrule
\end{tabular}
\caption{\textbf{Proportion of ground truth answers predicted by each heuristic}. The percentages show how many of the correct conclusions for a set of premises are predicted by the heuristics, for both valid and invalid syllogisms.}
\label{tab:heur_gt}
\end{table}

\begin{table*}
\begin{small}
\begin{center}
\begin{tabular}{cllll}
\toprule
\textbf{Syllogism} & \textbf{Atmosphere} & \textbf{Matching} & \textbf{Conversion} & \textbf{PHM} \\
\midrule
\multicolumn{5}{c}{Valid} \\

AA1 & Aac, Aca & Aac, Aca & Aac, Aca & Aac, Aca, Iac, Ica \\
AA2 & Aac, Aca & Aac, Aca & Aac, Aca & Aac, Aca, Iac, Ica \\
AA4 & Aac, Aca & Aac, Aca & Aac, Aca & Aac, Aca, Iac, Ica \\
AI2 & Iac, Ica & Iac, Ica, Oac, Oca & Iac, Ica & Ica, Oca \\
AI4 & Iac, Ica & Iac, Ica, Oac, Oca & Iac, Ica & Iac, Oac \\
AE1 & Eac, Eca & Eac, Eca & Eac, Eca & Eac, Oac \\
AE2 & Eac, Eca & Eac, Eca & Eac, Eca & Eca, Oca \\
AE3 & Eac, Eca & Eac, Eca & Eac, Eca & Eca, Oca \\
AE4 & Eac, Eca & Eac, Eca & Eac, Eca & Eac, Oac \\
AO3 & Oac, Oca & Iac, Ica, Oac, Oca & Oac, Oca & Oca, Ica \\
AO4 & Oac, Oca & Iac, Ica, Oac, Oca & Oac, Oca & Oac, Iac \\
IA1 & Iac, Ica & Iac, Ica, Oac, Oca & NVC & Iac, Oac \\
IA4 & Iac, Ica & Iac, Ica, Oac, Oca & NVC & Ica, Oca \\
IE1 & Oac, Oca & Eac, Eca & Oac, Oca & Eac, Oca \\
IE2 & Oac, Oca & Eac, Eca & Oac, Oca & Eca, Oca \\
IE3 & Oac, Oca & Eac, Eca & Oac, Oca & Eca, Oca \\
IE4 & Oac, Oca & Eac, Eca & Oac, Oca & Eac, Oac \\
EA1 & Eac, Eca & Eac, Eca & NVC & Eac, Oac \\
EA2 & Eac, Eca & Eac, Eca & NVC & Eca, Oca \\
EA3 & Eac, Eca & Eac, Eca & NVC & Eca, Oca \\
EA4 & Eac, Eca & Eac, Eca & NVC & Eac, Oac \\
EI1 & Oac, Oca & Eac, Eca & NVC & Eac, Oca \\
EI2 & Oac, Oca & Eac, Eca & NVC & Eac, Oac \\
EI3 & Oac, Oca & Eac, Eca & NVC & Eac, Oca \\
EI4 & Oac, Oca & Eac, Eca & NVC & Eac, Oac \\
OA3 & Oac, Oca & Iac, Ica, Oac, Oca & NVC & Eca, Oca \\
OA4 & Oac, Oca & Iac, Ica, Oac, Oca & NVC & Oac, Iac \\
\midrule
\multicolumn{5}{c}{Invalid} \\

AA3 & Aac, Aca & Aac, Aca & Aac, Aca & Aac, Aca, Iac, Ica \\
AI1 & Iac, Ica & Iac, Ica, Oac, Oca & Iac, Ica & Iac, Oac \\
AI3 & Iac, Ica & Iac, Ica, Oac, Oca & Iac, Ica & Ica, Oca \\
AO1 & Oac, Oca & Iac, Ica, Oac, Oca & Oac, Oca & Oac, Iac \\
AO2 & Oac, Oca & Iac, Ica, Oac, Oca & Oac, Oca & Oca, Ica \\
IA2 & Iac, Ica & Iac, Ica, Oac, Oca & NVC & Ica, Oca \\
IA3 & Iac, Ica & Iac, Ica, Oac, Oca & NVC & Iac, Oac \\
II1 & Iac, Ica & Iac, Ica, Oac, Oca & NVC & Iac, Ica, Oac, Oca \\
II2 & Iac, Ica & Iac, Ica, Oac, Oca & NVC & Iac, Ica, Oac, Oca \\
II3 & Iac, Ica & Iac, Ica, Oac, Oca & NVC & Iac, Ica, Oac, Oca \\
II4 & Iac, Ica & Iac, Ica, Oac, Oca & NVC & Iac, Ica, Oac, Oca \\
IO1 & Oac, Oca & Iac, Ica, Oac, Oca & NVC & Oac, Iac \\
IO2 & Oac, Oca & Iac, Ica, Oac, Oca & NVC & Oca, Ica \\
IO3 & Oac, Oca & Iac, Ica, Oac, Oca & NVC & Oca, Ica \\
IO4 & Oac, Oca & Iac, Ica, Oac, Oca & NVC & Oac, Iac \\
EE1 & Eac, Eca & Eac, Eca & NVC & Eac, Eca, Oac, Oca \\
EE2 & Eac, Eca & Eac, Eca & NVC & Eac, Eca, Oac, Oca \\
EE3 & Eac, Eca & Eac, Eca & NVC & Eac, Eca, Oac, Oca \\
EE4 & Eac, Eca & Eac, Eca & NVC & Eac, Eca, Oac, Oca \\
EO1 & Oac, Oca & Eac, Eca & NVC & Oac, Iac \\
EO2 & Oac, Oca & Eac, Eca & NVC & Oca, Ica \\
EO3 & Oac, Oca & Eac, Eca & NVC & Oca, Ica \\
EO4 & Oac, Oca & Eac, Eca & NVC & Oac, Iac \\
OA1 & Oac, Oca & Eac, Eca & NVC & Oac, Iac \\
OA2 & Oac, Oca & Eac, Eca & NVC & Oca, Ica \\
OI1 & Oac, Oca & Eac, Eca & NVC & Oca, Ica \\
OI2 & Oac, Oca & Eac, Eca & NVC & Oac, Iac \\
OI3 & Oac, Oca & Eac, Eca & NVC & Oca, Ica \\
OI4 & Oac, Oca & Eac, Eca & NVC & Oac, Iac \\
OE1 & Oac, Oca & Eac, Eca & NVC & Oca, Ica \\
OE2 & Oac, Oca & Eac, Eca & NVC & Oca, Ica \\
OE3 & Oac, Oca & Eac, Eca & NVC & Oac, Iac \\
OE4 & Oac, Oca & Eac, Eca & NVC & Oca, Ica \\
OO1 & Oac, Oca & Eac, Eca & NVC & Oac, Iac \\
OO2 & Oac, Oca & Eac, Eca & NVC & Oca, Ica \\
OO3 & Oac, Oca & Eac, Eca & NVC & Oca, Ica \\
OO4 & Oac, Oca & Eac, Eca & NVC & Oac, Oca, Iac, Ica \\
\bottomrule
\end{tabular}
\end{center}
\end{small}
\caption{\textbf{Predicted conclusions for all heuristic theories.} Table adapted from \citet{khem:theo12}.}\label{tab:heur_preds}
\end{table*}

\section{Example Mistakes} 
\label{app:examples}

We show in Figure \ref{fig:in_error} a mistake done by the LLaMA-3 8B ICL$_{in}$. Even though the model is shown examples of the same schema where the given answer is \textit{nothing follows}, the model still generates a different conclusion.

Instead, we show in Figure \ref{fig:cot_error} a mistake by LLaMA-3 8B in the ZS-CoT setting. The correct conclusion in this example would be \textit{nothing follows} but the model produces three wrong ones.  

\begin{figure*}[t]
\begin{center}
\begin{tikzpicture}
\node[draw, rounded corners, text width=.9\linewidth, align=flush left, inner sep=16 pt] (box)
{
\small

\textbf{INPUT:}

You will be presented with premises together with eight possible conclusions and the option 'Nothing follows'. Write the conclusion that logically follows given the premises or 'Nothing follows' if none of the other conclusions logically follow. Read the passage of information thoroughly and select the correct answer from the available options. Read the premises thoroughly to ensure you know what the premise entails.

\vspace{4pt}
\sethlcolor{warmcream}
\hl{
Use the examples in the following context to better answer the test examples:

Syllogism:

Premise 1: All raiahl are mcaiamp.
Premise 2: All meok are mcaiamp.

<shuffled-options>

Answer: Nothing follows.

\textbf{...}

Syllogism:

Premise 1: All flioungs are theauft.
Premise 2: All schwiisly are theauft.

<shuffled-options>

Answer: Nothing follows.}

\vspace{4pt}
Complete the following test example:

Syllogism:

Premise 1: All siameses are felines.
Premise 2: All cats are felines.

<shuffled-options>

Answer:  Given the premises I choose the following option(s):

\vspace{10pt}
\textbf{GENERATED TEXT:}

\sethlcolor{pink}
\hl{Some cats are siameses.}
};
\end{tikzpicture}
\end{center}
\caption{\textbf{ICL$_{in}$ Mistake.} The example is taken from LLaMA-3 8B. In the above example, the nine options for each in-context example are substituted with the special token \texttt{<shuffled-options>} for ease of presentation. The \sethlcolor{warmcream}\hl{in-context examples} are highlighted.}\label{fig:in_error}
\end{figure*}

\begin{figure*}[t]
\begin{center}
\begin{tikzpicture}
\node[draw, rounded corners, text width=.9\linewidth, align=flush left, inner sep=16 pt] (box)
{
\small

\textbf{INPUT:}

You will be presented with premises together with eight possible conclusions and the option 'Nothing follows'. Write the conclusion that logically follows given the premises or 'Nothing follows' if none of the other conclusions logically follow. Read the passage of information thoroughly and select the correct answer from the available options. Read the premises thoroughly to ensure you know what the premise entails.

\vspace{4pt}
Complete the following test example:

Syllogism:

Premise 1: All chickadees are winged animals.
Premise 2: All birds are winged animals.

Options:
Some birds are not chickadees.
No chickadees are birds.
Some birds are chickadees.
All birds are chickadees.
Some chickadees are birds.
Some chickadees are not birds.
Nothing follows.
All chickadees are birds.
No birds are chickadees.

Answer:  Let's think this through, step by step.

\vspace{10pt}
\textbf{GENERATED TEXT:}

\sethlcolor{pink}
\hl{  We know that all chickadees are winged animals.  We also know that all birds are winged animals.  So, we know that all chickadees are winged animals and all birds are winged animals. So, my final answer(s) is/are:  All chickadees are birds.  All birds are chickadees.  Some birds are chickadees.}
};
\end{tikzpicture}
\end{center}
\caption{\textbf{ZS-CoT Mistake.} The example is taken from LLaMA-3 8B. In the above example, the nine options for each in-context example are substituted with the special token \texttt{<shuffled-options>} for ease of presentation.}\label{fig:cot_error}
\end{figure*}

\end{document}